\documentclass{article}
\usepackage{iclr2027_conference,times}

\usepackage{amsmath,amsfonts,bm}

\def\eqref#1{equation~\ref{#1}}
\def\1{\bm{1}}

\DeclareMathAlphabet{\mathsfit}{\encodingdefault}{\sfdefault}{m}{sl}
\SetMathAlphabet{\mathsfit}{bold}{\encodingdefault}{\sfdefault}{bx}{n}

\usepackage{hyperref}
\usepackage{url}
\usepackage{xspace}
\usepackage{enumitem}
\usepackage{graphicx}
\usepackage{wrapfig}
\usepackage{booktabs}
\usepackage{multirow}
\usepackage{tabularx}
\usepackage{makecell}
\usepackage{needspace}
\usepackage{colortbl}
\usepackage{xcolor}
\usepackage{array}
\usepackage{float}
\newcommand{\method}{\textsc{DrafTS}\xspace}
\hypersetup{
    colorlinks=true,
    citecolor=blue, % 文献引用
    linkcolor=blue, % 图、表、公式等交叉引用
    urlcolor=blue   % 网址
}

\title{\raggedright
DrafTS: Time-Aware Decomposition with Residual Correction
for Time Series Modeling}

\iclrfinalcopy
\author{
Yiqiu Liu$^1$, Siru Zhong$^1$, Zhiguang Wang$^2$, Qingsong Wen$^3$, Yuxuan Liang$^1$\\[3pt]
{\normalfont\small $^1$The Hong Kong University of Science and Technology (Guangzhou), China}\\
{\normalfont\small $^2$Abel AI Lab, USA \quad $^3$Squirrel Ai Learning, USA}
}

\begin{document}

\maketitle

\fancyhead{}
\renewcommand{\headrulewidth}{0pt}
\setlength{\headsep}{12pt}
\raggedbottom
% \vspace{-0.15in}

\begin{abstract}
Real-world time series contain evolving underlying dynamics with irregular variations that lack stable temporal patterns and are often referred to as noise. Existing methods address this mixture by filtering frequencies or suppressing noisy observations. They either miss temporal evolution or risk suppressing useful dynamics. We propose \textbf{\method}, a model-agnostic framework that aims to reduce noise while preserving evolving dynamics through time-aware \underline{D}ecomposition with \underline{R}esidu\underline{A}l correction \underline{F}or \underline{T}ime \underline{S}eries. \method uses features derived from instantaneous amplitude and frequency to guide decomposition into a primary component intended to capture underlying dynamics. A task-specific backbone models the primary component, while a lightweight correction module uses residual information to correct the backbone output. Across four time series modeling tasks, \method  improves six diverse backbones, demonstrating its effectiveness. Code is at \url{https://github.com/Autumn61q/DrafTS}.

% 匿名代码库：https://anonymous.4open.science/r/DrafTS-A14F/

\end{abstract}

\section{Introduction}
\label{intro}

Deep learning is widely used for time series modeling, supporting tasks such as forecasting \citep{pangu}, classification \citep{kiyasseh2021clocs} and anomaly detection \citep{xu2022anomaly}. Recent recurrent \citep{Lai2017ModelingLA}, convolutional \citep{timesnet}, MLP-based \citep{timemixer, dlinear}, and attention-based \citep{crossformer,itransformer} architectures have improved temporal dependency modeling. Yet these improvements do not necessarily enable models to distinguish underlying dynamics from noise\footnote{Throughout this paper, underlying dynamics refer to structured temporal patterns, such as trends and periodic behavior, whereas noise refers to irregular variations that do not follow stable temporal patterns.}. This distinction becomes harder in real-world nonlinear and nonstationary time series, where the dynamics themselves evolve over time \citep{emd}. Reducing noise while preserving these evolving dynamics therefore remains challenging.

\begin{figure}[!t]
    \vspace{-10pt}
    \centering \includegraphics[width=\linewidth]{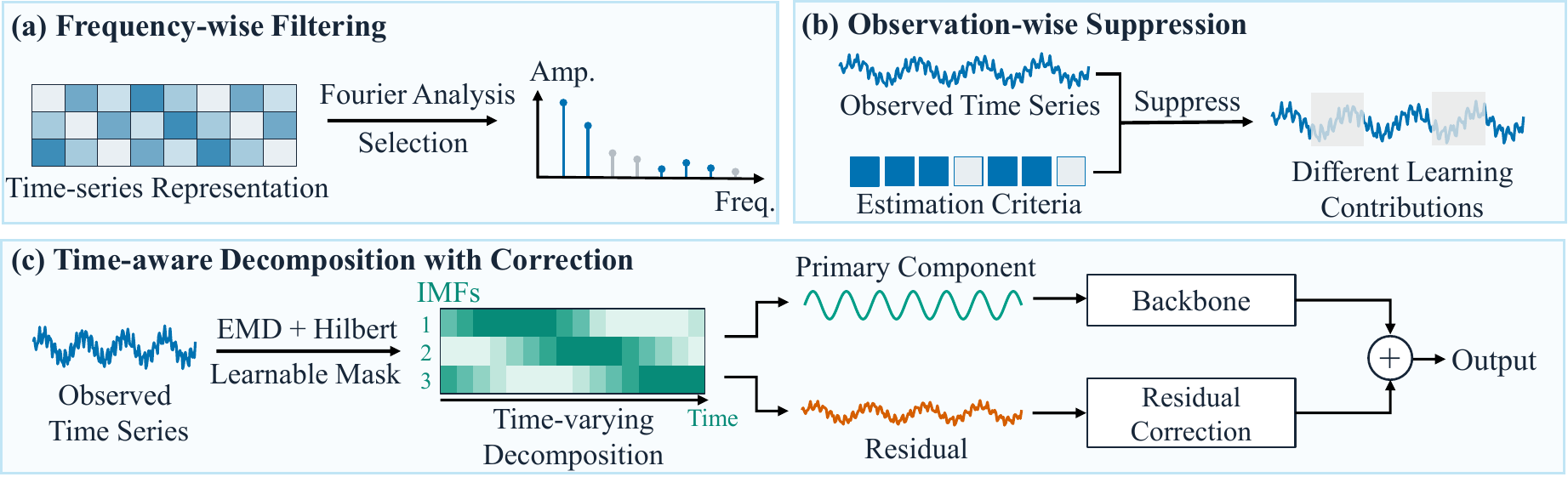}
    \vspace{-15pt}
    \caption{
    \textbf{Modeling paradigms for underlying dynamics and noise.}
    \textit{(a) Frequency-wise filtering} overlooks temporal evolution.
    \textit{(b) Observation-wise suppression} limits learning from noisy observations, suppressing useful dynamics within them.
    \textit{(c) Time-Aware Decomposition with Correction} tracks evolving dynamics and exploits dynamics left in the residual.
}
    \label{fig:intro}
    \vspace{-10pt}
\end{figure}

Existing deep learning methods pursue this balance mainly via two paradigms (Figure~\ref{fig:intro} (a,b)). The first is \textbf{frequency-wise filtering}, which selects spectral components in time-series representations. FITS \citep{fits} uses a fixed frequency cutoff, while FiLM \citep{film} and FilterNet \citep{filternet} retain selected Fourier representations. However, such global filtering cannot track the temporal evolution of the time series. The second is \textbf{observation-wise suppression}, which uses noise estimates to regulate learning in the time domain. RobustTSF \citep{robusttsf} discards anomalous windows, Selective Learning \citep{selectivelearning} masks uncertainty or anomaly in the loss and DropoutTS \citep{dropoutts} adapts instance-level dropout. However, suppressing noisy observations as a whole may also hinder learning useful dynamics within them. Moreover, both paradigms may reduce the impact of the noise at the expense of useful dynamics. 

Decomposition offers a possible way to distinguish evolving dynamics from noise. However, commonly used Fourier and discrete cosine transforms \citep{fft,dct} apply fixed-frequency bases, which fail to track the temporal evolution of nonlinear, nonstationary series. As shown in Figure~\ref{fig:intro} (c), we propose a third paradigm: \textbf{Time-aware decomposition with correction}. Instead of frequency filtering or observation-wise suppression, we extract a primary component to capture evolving dynamics and recover missed dynamics from the residual.

To this end, we propose \method, a model-agnostic framework that uses time-aware \textbf{D}ecomposition to retain evolving dynamics and corrects backbone predictions with \textbf{R}esidu\textbf{A}l information \textbf{F}or \textbf{T}ime \textbf{S}eries modeling. Specifically, \textbf{in the decomposition stage}, we track time-varying oscillations using instantaneous frequency obtained through the Hilbert transform \citep{gabor1946theory,boashash1992instantaneous}. Since instantaneous frequency can be difficult to interpret when fast and slow oscillations coexist, we first apply empirical mode decomposition (EMD) to decompose the input into intrinsic mode functions (IMFs) without predefined bases \citep{emd}. 
\begin{wrapfigure}{r}{0.64\textwidth}
    \centering
    \vspace{-0.5\baselineskip}
    \includegraphics[width=\linewidth]{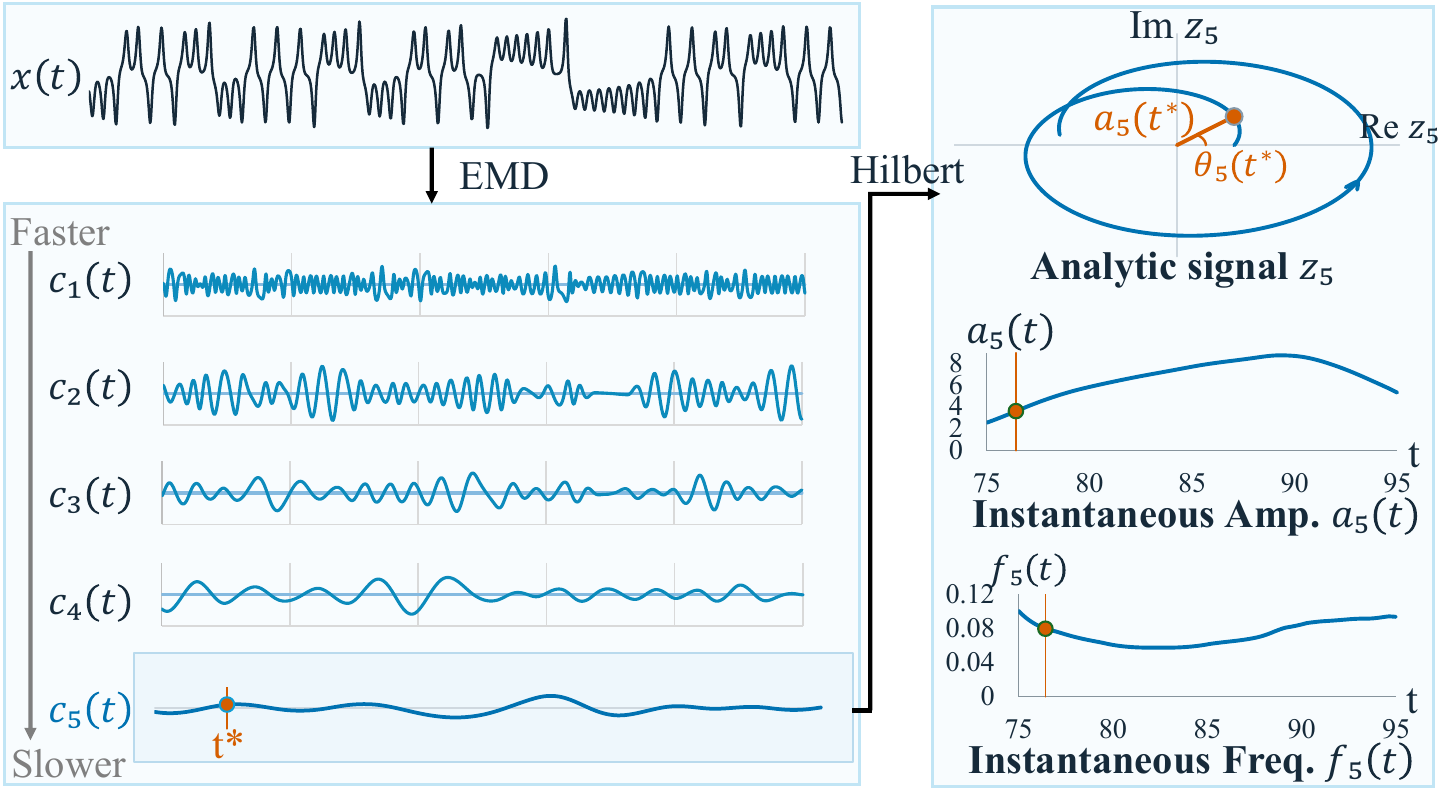}
    \vspace{-15pt}
    \caption{\textbf{EMD and Hilbert transform example.} The first five IMFs are shown. Applying the Hilbert transform to $c_5(t)$ yields $z_5(t)$ and its instantaneous amplitude and frequency. Re and Im denote real and imaginary parts; orange markers mark time $t^*$.}
    \label{fig:emd_hilbert_demo}
    \vspace{-0.6\baselineskip}
\end{wrapfigure}
Secondly, we apply the Hilbert transform to each IMF to construct its analytic signal, from which we derive its instantaneous amplitude and frequency. Figure~\ref{fig:emd_hilbert_demo} illustrates this process. Features derived from these quantities guide a time-aware mask that decomposes the input into a primary component and a residual.
\textbf{In the modeling stage}, a task-specific backbone models the primary component, while a lightweight \textit{Residual-Aware Correction} module exploits useful dynamics remaining in the residual to correct the backbone output. The module processes the residual sequence recurrently with fixed hidden weights, producing nonlinear representations; only the output layer is trained, so the module introduces few trainable parameters. Finally, we combine the residual correction with the backbone output to produce the task prediction. In summary, our contributions are as follows:
% \vspace{-0.5em} 盲审
\vspace{-0.5em}
\begin{itemize}[leftmargin=*]

    \item \textbf{Novel Paradigm:} We introduce \textit{Time-Aware Decomposition with Correction}. \method uses time-aware masks to capture evolving dynamics in a primary component and exploits useful dynamics left in the residual by imperfect decomposition to correct the backbone output.
    
    % \vspace{-0.0em} 盲审

    \item \textbf{Universal Compatibility:} \method augments diverse task-specific backbones with time-aware decomposition and lightweight residual correction, without modifying their internal architectures.

    % \vspace{-0.0em} 盲审

    \item \textbf{Superior Performance:} Across four time series modeling tasks, \method improves diverse backbones, demonstrating its generality and effectiveness.
    
\end{itemize}

\section{Related Work}
\label{sec:relatedwork}

\paragraph{Deep Learning for Time Series Modeling.}
Learning temporal dynamics that support accurate prediction is a central objective of deep learning for time series modeling. Earlier approaches use patches \citep{informer,patchtst}, dependency modeling \citep{crossformer,itransformer}, linear prediction \citep{dlinear}, or multiscale mixing \citep{timemixer}. More recently, TimeFilter and L-Drive \citep{timefilter,ldrive} refine dependency and change modeling, while large time series models extend pre-training \citep{chronos,moirai}, long contexts \citep{timerxl}, and sparse experts \citep{timemoe}. However, stronger modeling capacity alone does not ensure that evolving dynamics are distinguished from noise.

\paragraph{Modeling paradigms for noise.}
Prior work limits noise effects via spectral filtering or regulating observation use. \textbf{Frequency-wise filtering} includes fixed cutoffs in FITS \citep{fits} and Fourier selection in FiLM \citep{film} and FilterNet \citep{filternet}. However, global spectral operations fail to reveal how oscillations change within a window \citep{fcvae}. \textbf{Observation-wise suppression} instead adjusts learning according to estimated noise, through window rejection in RobustTSF \citep{robusttsf}, loss masking of uncertain or anomalous timesteps in Selective Learning \citep{selectivelearning}, and instance-adaptive dropout in DropoutTS \citep{dropoutts}. Yet rejecting or masking noisy observations may limit learning useful dynamics within them.

\paragraph{Signal Decomposition Methods.}
Autoformer \citep{autoformer} and FEDformer \citep{fedformer} use moving averages for trend--seasonal decomposition. As they do not aim to separate evolving dynamics from noise, we do not discuss them further and instead focus on methods relevant to this goal. Fourier \citep{fft} and discrete cosine transforms (DCT) \citep{dct} represent each window as a linear combination of fixed-frequency bases. For nonlinear and nonstationary data, global stationary bases can yield physically ambiguous frequencies. Short-time Fourier transform (STFT) \citep{stft} mitigates this issue with shorter windows, but fixed windows still limit tracking evolving oscillations. In contrast, EMD extracts time-domain oscillatory modes \citep{emd}. With the Hilbert transform, it yields interpretable instantaneous amplitude and frequency, and is used for denoising across domains \citep{hassan2005empirical,quilfen2021denoising,quilfen2022towards}. However, threshold-based EMD denoising \citep{emd_denoise_2009} may remove underlying dynamics. \method instead learns task-supervised, time-aware soft masks from analytic features and keeps the residual for correction.  

\section{Methodology}
\label{sec:methodology}

\subsection{Notations and Problem Formulations}
Let $\mathbf{X}=[\mathbf{x}_1,\ldots,\mathbf{x}_T]^\top\in\mathbb{R}^{T\times N}$ be a multivariate series with $T$ time steps and $N$ variables. For $n=1,\ldots,N$, let $x_{1:T}^{(n)}=(x_1^{(n)},\ldots,x_T^{(n)})$ denote the sequence of the $n$-th variable. Let $\mathbf{Y}$ denote the task-specific target. For forecasting, $\mathbf{Y}\in\mathbb{R}^{S\times N}$ contains the subsequent $S$ time steps. For classification, $\mathbf{Y}\in\{0,1\}^{C}$ is the one-hot label vector, where $C$ is the number of classes. For anomaly detection, $\mathbf{Y}$ is the next observation $\mathbf{x}_{t+1}\in\mathbb{R}^{N}$.

\begin{figure}[!t]
    % \vspace{-10pt}
    \centering \includegraphics[width=\linewidth]{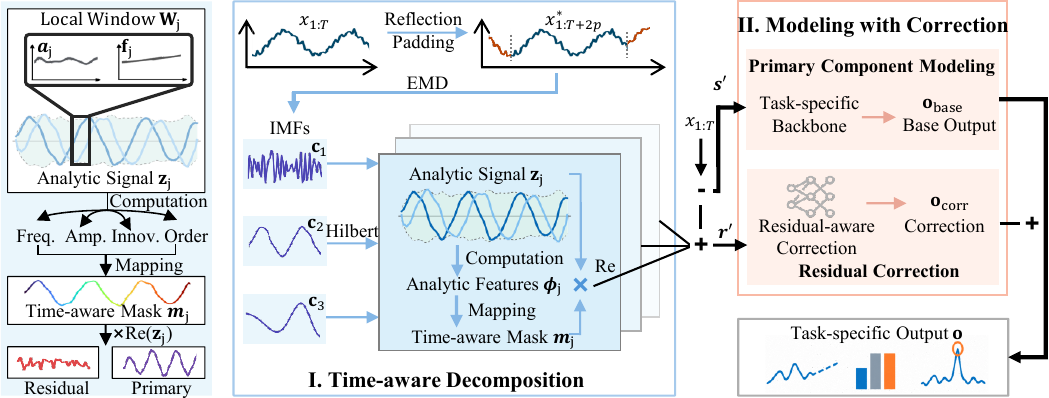}
    \vspace{-15pt}
    \caption{\textbf{Overview of \method.}
    \textbf{Right:} Time-aware decomposition splits $x_{1:T}$ into the primary component $\mathbf{s}'$ and residual $\mathbf{r}'$. The backbone and \textit{Residual-Aware Correction} (RC) produce the base output and residual correction, respectively, which are summed to obtain $\mathbf{o}$. \textbf{Left:} Enlarged view of the blue shaded box showing how analytic features are mapped to time-aware masks for each IMF.}
    \vspace{-10pt}
    \label{fig:framework}
\end{figure}

Figure~\ref{fig:framework} illustrates the overall framework of \method. Our method augments a task-specific backbone through two stages. The decomposition stage splits $\mathbf{X}=\mathbf{S}'+\mathbf{R}'$ into a primary component $\mathbf{S}'$ and the residual $\mathbf{R}'$, both in $\mathbb{R}^{T\times N}$. The modeling stage uses a task-specific backbone to model $\mathbf{S}'$ and a fixed reservoir with a trainable output layer to derive an output correction from $\mathbf{R}'$.

\subsection{Stage I: Time-aware Decomposition}

\paragraph{EMD and the Hilbert transform.}
We apply EMD independently to each variable $x_{1:T}^{(n)}$. We omit $n$ for clarity. For $x_{1:T}$, EMD extracts oscillatory modes by iterative sifting \citep{emd}.
To extract the first IMF, set $h_t^{(0)} \leftarrow x_t$. At the $q$-th sifting step, EMD locates local extrema of $h_t^{(q)}$. Interpolating local maxima and minima gives the upper and lower envelopes $e_t^{+,(q)}$ and $e_t^{-,(q)}$, respectively. Their pointwise average defines the local envelope mean $\mu_t^{(q)}=\frac{e_t^{+,(q)}+e_t^{-,(q)}}{2}.$ The local envelope mean is then removed by $h_t^{(q+1)}=h_t^{(q)}-\mu_t^{(q)}.$ Once $h_t^{(q)}$ satisfies the IMF conditions, it is recorded as $c_{j,t}$ and removed from the current signal. Repeating this process for $J$ modes gives
\begin{equation*}
    x_t
    =
    \sum_{j=1}^{J}c_{j,t}
    +
    \rho_t^{(J)},
\end{equation*}
where $\rho_t^{(J)}$ is the component remaining after the first $J$ IMFs
have been extracted. Generally, earlier IMFs capture higher-frequency oscillations, while later IMFs capture lower-frequency ones.

For each IMF $\mathbf{c}_j\in\mathbb{R}^{T}$, the Hilbert transform uniquely specifies its analytic signal $z_{j,t}=c_{j,t}+\mathrm{i}\mathcal{H}[\mathbf{c}_j]_t=a_{j,t}e^{\mathrm{i}\theta_{j,t}}$, whose instantaneous amplitude, phase, and frequency are defined as
\begin{equation}
a_{j,t}=|z_{j,t}|,
\quad
\theta_{j,t}=\operatorname{unwrap}\!\bigl(\arg z_{j,t}\bigr),
\quad
f_{j,t}
=
\frac{\theta_{j,t}-\theta_{j,t-1}}{2\pi\Delta t}.
\end{equation}

Here, $\operatorname{unwrap}$ removes $2\pi$ phase jumps along the time axis. Together, EMD and the Hilbert transform provide a physically interpretable representation of each oscillatory mode without assuming linearity or stationarity. We therefore use them as the basis for the following decomposition. 

\paragraph{EMD for padded sequence.}
EMD is sensitive to boundary effects because no extrema are available beyond the observed window. This missing information can distort the interpolated envelopes and the resulting IMFs. Therefore, to obtain reliable IMFs, we apply reflection padding for the time series. Specifically, we extend both ends of the time series by reflecting the nearest $p$ observations:
\begin{equation*}
x^*_{1:T+2p}
=
[
x_{p+1},\ldots,x_2,\,
x_1,\ldots,x_T,\,
x_{T-1},\ldots,x_{T-p}
].
\end{equation*}
The reflected samples provide symmetric support beyond each boundary.
Consequently, the envelopes near the original endpoints are estimated from extrema on both sides rather than through one-sided extrapolation, reducing boundary distortion in the resulting IMFs. We empirically set the padding length to one quarter of the input length, i.e., $p=0.25T$.

We apply EMD to the padded sequence $x^*_{1:T+2p}$. Then we discard the padding and use the central $T$ timesteps of each IMF, denoted by $c_{j,t}$, $t=1,\ldots,T$, in subsequent processing. Previous application studies have observed noise mainly in the first two or three IMFs \citep{hassan2005empirical,quilfen2021denoising,quilfen2022towards}. Therefore, we examine the first $J\in\{2,3\}$ IMFs for potential noise. The following local analysis determines their primary component and the residual at each timestep.

\paragraph{Analytic representation and feature setup.}

For each IMF $\mathbf{c}_j$, $j=1,\ldots,J$, the Hilbert transform yields its analytic signal. We define the local window over which the features are computed and an amplitude-based weight that controls the contribution of each transition:

\begin{itemize}[leftmargin=*, labelsep=0.5em]

\item \textit{Local window $\mathcal{W}_{j,\tau}$:}
Let $g_{j,u}=|\nabla_u\theta_{j,u}|$ denote the absolute phase
gradient, where $u=1,\ldots,T$ indexes samples and $\nabla_u$
uses unit sample spacing. We estimate the oscillation rate of $\mathbf{c}_j$ as $\tilde f_j=\operatorname{median}_{u:g_{j,u}>\epsilon}(g_{j,u})/(2\pi)$ in cycles/sample and set $L_j=\operatorname{round}(1.5/\tilde f_j)$. At timestep $\tau$, we define the local window $\mathcal{W}_{j,\tau}=[\tau-L_j/2,\tau+L_j/2]$ and compute the subsequent statistical features within it. This design balances statistical stability and temporal locality.

\item \textit{Amplitude-based weight $w_{j,\tau}$:}
Because phase is sensitive to perturbations at low amplitude, we weight the transition from $\tau-1$ to $\tau$ by adjacent amplitudes $a_{j,\tau}a_{j,\tau-1}$:
\begin{equation}w_{j,\tau}=\min\left\{1,\frac{a_{j,\tau}a_{j,\tau-1}}{\left(\operatorname{median}_{u=1,\ldots,T}(a_{j,u})\right)^2}\right\}.\end{equation}
The squared median normalizes each IMF's weights by its amplitude scale; clipping at $1$ prevents high-amplitude amplification. Low-amplitude transitions thus receive smaller weights.

\end{itemize}

\paragraph{Four analytic features.}
We compute four analytic features to guide the time-aware mask. The first three characterize local temporal variation and are illustrated in Figure~\ref{fig:lorenz-qai}.

\begin{enumerate}[leftmargin=*, labelsep=0.5em]

    \item \textbf{Frequency variation ($F_{j,t}$):} Measures the magnitude of instantaneous frequency changes:
    \begin{equation}F_{j,t}
    =
    \frac{
    \sqrt{\sum_{\tau\in\mathcal W_{j,t}}
    \bar w_{j,t,\tau}(\Delta f_{j,\tau})^2}
    }{
    \sum_{\tau\in\mathcal W_{j,t}}
    \bar w_{j,t,\tau}|f_{j,\tau}|
    },
    \qquad
    \Delta f_{j,\tau}=f_{j,\tau}-f_{j,\tau-1},
    \end{equation} 
     where $\bar w_{j,t,\tau}=\frac{w_{j,\tau}}{\sum_{u\in\mathcal W_{j,t}}w_{j,u}}$ is the normalized weight for $\tau\in\mathcal W_{j,t}$. The numerator emphasizes large frequency changes through squared increments. As a second difference of phase, $\Delta f_{j,\tau}$ is sensitive to perturbations, so we use the local weighted mean of $|f_j|$ for stable normalization.

    \item \textbf{Amplitude variation ($A_{j,t}$):} Measures the magnitude of instantaneous amplitude changes: 
    \begin{equation}
        A_{j,t}=\sqrt{\sum_{\tau\in\mathcal W_{j,t}}\bar w_{j,t,\tau}(\Delta a_{j,\tau})^2},\qquad\Delta a_{j,\tau}=\frac{2|a_{j,\tau}-a_{j,\tau-1}|}{a_{j,\tau}+a_{j,\tau-1}},
    \end{equation}
    Similar to $F_{j,t}$, the numerator emphasizes large amplitude changes. Compared with $F_{j,t}$, $A_{j,t}$ directly measures the first-order change between adjacent amplitudes. Therefore, the two nonnegative amplitudes provide a symmetric local reference for normalizing this change.

    \item \textbf{Analytic innovation ($I_{j,t}$):} Measures the instantaneous evolution that cannot be captured by a local transition coefficient. We estimate this coefficient by minimizing the weighted one-step approximation error, whose closed-form solution is
    \begin{equation}\alpha_{j,t}
    =
    \frac{
    \sum_{\tau\in\mathcal W_{j,t}}
    \bar w_{j,t,\tau}z_{j,\tau}z_{j,\tau-1}^{*}
    }{
    \sum_{\tau\in\mathcal W_{j,t}}
    \bar w_{j,t,\tau}|z_{j,\tau-1}|^2
    }.\end{equation} 
    Writing $\alpha_{j,t}=\rho_{j,t}e^{\mathrm{i}\psi_{j,t}}$, we have $\alpha_{j,t}z_{j,\tau-1}=\rho_{j,t}a_{j,\tau-1}e^{\mathrm{i}(\theta_{j,\tau-1}+\psi_{j,t})}.$ This expression estimates $z_{j,\tau}$ by scaling the previous amplitude by $\rho_{j,t}$ and advancing the previous phase by $\psi_{j,t}$. 
    We then define the analytic innovation as the normalized error remaining after applying $\alpha_{j, t}$:
    \begin{equation} I_{j,t}
    =
    \frac{
    \sum_{\tau\in\mathcal W_{j,t}}
    \bar w_{j,t,\tau}
    |z_{j,\tau}-\alpha_{j,t}z_{j,\tau-1}|^2
    }{
    \sum_{\tau\in\mathcal W_{j,t}}
    \bar w_{j,t,\tau}|z_{j,\tau}|^2
    }.
    \end{equation}
    It measures the instantaneous evolution remaining unexplained after estimating $z_{j,\tau}$ from $z_{j,\tau-1}$ using $\alpha_{j,t}$. A small $I_{j,t}$ indicates consistent evolution across adjacent timesteps.
    
    \item \textbf{IMF order ($O_j$):} Encodes the normalized order index of the $j$-th IMF among $J$ IMFs:
    \begin{equation}
    O_j=1-\frac{j-1}{\max(J-1,1)},
    \qquad j=1,\ldots,J.
    \end{equation} 
    $O_j$ lets the model distinguish IMFs by their high-to-low frequency order.

\end{enumerate}

\paragraph{Time-aware mask and decomposition.}
We concatenate the four features as $\boldsymbol{\phi}_{j,t}=[F_{j,t},A_{j,t},I_{j,t},O_{j}]^\top \in\mathbb R^4$ and map them to a time-aware mask:
\begin{equation}
\label{eq:time-varying maks}
m_{j,t}=m_{\max}\sigma\left((\mathbf{v}^{\top}\boldsymbol{\phi}_{j,t}+b)\right),
\end{equation}
where $\mathbf{v}\in\mathbb{R}^4$ and $b\in\mathbb{R}$ are learnable, $\sigma$ is the sigmoid function, and $m_{\max}$ sets the maximum fraction assigned to the residual. Each candidate IMF contributes $m_{j,t}c_{j,t}$ to the residual, yielding
\begin{equation}
r'_t=\sum_{j=1}^{J}m_{j,t}c_{j,t},\qquad s'_t=x_t-r'_t,
\end{equation}
where $s'_t$ is intended to capture underlying dynamics and $r'_t$ is the residual. Applying this decomposition independently to all $N$ variables yields $\mathbf{R}',\mathbf{S}'\in\mathbb{R}^{T\times N}$.

\subsection{Stage II: Modeling with Correction}

\paragraph{Residual-Aware Correction (RC) with a fixed reservoir.} 
A recurrent reservoir with input weights $\mathbf{W}_{\mathrm{in}}$ and recurrent weights $\mathbf{W}_{\mathrm{res}}$ maps sequence $\mathbf{u}_{1:T}$ to output $\mathbf{y}$ via recurrent states $\mathbf{h}_t$:
\begin{equation}
\label{eq:fixed_reservoir}
\mathbf{h}_t=\tanh\left(\mathbf{W}_{\mathrm{in}}\mathbf{u}_t+\mathbf{W}_{\mathrm{res}}\mathbf{h}_{t-1}\right),
\qquad
\mathbf{y}=\mathbf{W}_{\mathrm{out}}\mathbf{h}_T+\mathbf{b}_{\mathrm{out}}.
\end{equation}
Prior work shows that the reservoir can remain fixed \citep{jaeger2001echo,maass2002realtime}.
With suitably chosen heterogeneous initializations, the reservoir maps input history into diverse nonlinear states, providing an expressive temporal representation. Only the output layer $[\mathbf{W}_{\mathrm{out}}, \mathbf{b}_{\mathrm{out}}]$ is trained. 

The residual $\mathbf{R}'$ may contain limited underlying dynamics that require expressive nonlinear temporal modeling. However, using a heavily parameterized model to capture these limited dynamics would be inefficient. We therefore build our \textit{Residual-Aware Correction (RC)} module on the fixed recurrent reservoir with a task-specific output layer. Specifically, we use the residual as the reservoir input, $\mathbf{u}_t\leftarrow\mathbf{R}'_{t,:}$, and use the output as the task-specific correction, i.e., $\mathbf{y}\leftarrow\mathbf{o}_{\mathrm{corr}}$. Eventually, we obtain expressive temporal representations with few trainable parameters. 

\paragraph{Backbone modeling with residual correction.}
As shown in Figure~\ref{fig:framework}, the task-specific backbone maps the primary component $\mathbf{S}'$ to the base output $\mathbf{o}_{\mathrm{base}}$, while RC maps the residual $\mathbf{R}'$ to a correction $\mathbf{o}_{\mathrm{corr}}$. The two outputs are then combined to produce the final task output:
\begin{equation}
\mathbf{o}=\mathbf{o}_{\mathrm{base}}+\mathbf{o}_{\mathrm{corr}}.
\end{equation}
For forecasting, $\mathbf{o}\in\mathbb{R}^{S\times N}$ predicts the future sequence $\mathbf{Y}$. For classification, $\mathbf{o}\in\mathbb{R}^{C}$ gives class logits. Applying softmax yields class probabilities. For anomaly detection, $\mathbf{o}\in\mathbb{R}^{N}$ gives next-step prediction whose error defines the anomaly score.

\paragraph{Loss functions.}
The overall training objective of \method is: 
\begin{equation}
\begin{aligned}
\mathcal{L}
&=
\mathcal{L}_{\mathrm{base}}
+\lambda_1\mathcal{L}_{\mathrm{corr}}
+\lambda_2\mathcal{L}_{\mathrm{fused}}
\\
&=
\ell_{\mathrm{task}}(\mathbf{o}_{\mathrm{base}},\mathbf{Y})
+\lambda_1\ell_{\mathrm{task}}\left(
\operatorname{detach}(\mathbf{o}_{\mathrm{base}})
+\mathbf{o}_{\mathrm{corr}},\mathbf{Y}
\right)
+\lambda_2\ell_{\mathrm{task}}\left(
\mathbf{o}_{\mathrm{base}}+\mathbf{o}_{\mathrm{corr}},\mathbf{Y}
\right).
\end{aligned}
\end{equation}

Short- and long-term forecasting use sMAPE and MAE/MSE as $\ell_{\mathrm{task}}$, respectively; classification and anomaly detection use cross-entropy and one-step MSE. $\mathcal{L}_{\mathrm{base}}$ trains the backbone and mask via $\mathbf{S}'$, $\mathcal{L}_{\mathrm{corr}}$ trains RC on the residual with $\mathbf{o}_{\mathrm{base}}$ detached, and $\mathcal{L}_{\mathrm{fused}}$ supervises the fused output.
\section{Experiments}

To verify the effectiveness of \method, we perform extensive experiments across four mainstream modeling tasks, including short- and long-term forecasting, classification and anomaly detection.

\subsection{Experimental Settings}

% \Needspace{0.42\textheight}
% \begin{wraptable}{r}{0.68\textwidth}
\begin{table}[thbp]
\small
\centering
\vspace{-15pt}
\caption{\textbf{Summary of the experiment datasets.}
Further details are provided in Appendix~\ref{app:benchmark_details}.}
\label{tab:exp_settings}
\setlength{\tabcolsep}{7pt} % 0.5相邻两列之间的实际间距
\renewcommand{\arraystretch}{1.0} % 行高
\renewcommand{\tabularxcolumn}[1]{m{#1}}
\newcolumntype{C}[1]{>{\centering\arraybackslash}m{#1}}
\newcolumntype{L}[1]{>{\raggedright\arraybackslash}m{#1}}

\resizebox{1\columnwidth}{!}{
\begin{tabular}{@{}c|L{0.35\textwidth}|C{0.26\textwidth}|C{0.15\textwidth}@{}}
\toprule
\textbf{Tasks}
& \multicolumn{1}{c|}{\textbf{Datasets}}
& \textbf{Metrics}
& \textbf{Series Length} \\
\midrule

\multirow[c]{2}{*}[-4pt]{Forecasting}
& \textbf{Long-term:} ETTh1, ETTm1, Weather, Electricity, ILI,
ECG, Laser, Noisy Lorenz63 (SNR=\{3,5,7\})
& MSE, MAE
& $16\sim720$ \\[2pt]

\cmidrule(l{4pt}r{4pt}){2-4}

& \textbf{Short-term:} M4 (6 subsets)
& sMAPE, MASE, OWA
& $6\sim48$ \\[2pt]

\midrule
Classification
& UEA (10 subsets)
& Accuracy
& $29\sim1152$ \\

\midrule
Anomaly Detection
& SMD, MSL, SMAP, SWaT, PSM
& Precision, Recall, F1-Score
& 100 \\

\bottomrule
\end{tabular}
}
\vspace{-10pt}
\end{table}
% \vspace{-8pt}
% \end{wraptable} 

\paragraph{Datasets.}

\begin{wrapfigure}{r}{0.51\columnwidth}
  \vspace{-0pt}
  \centering
  \includegraphics[width=\linewidth]{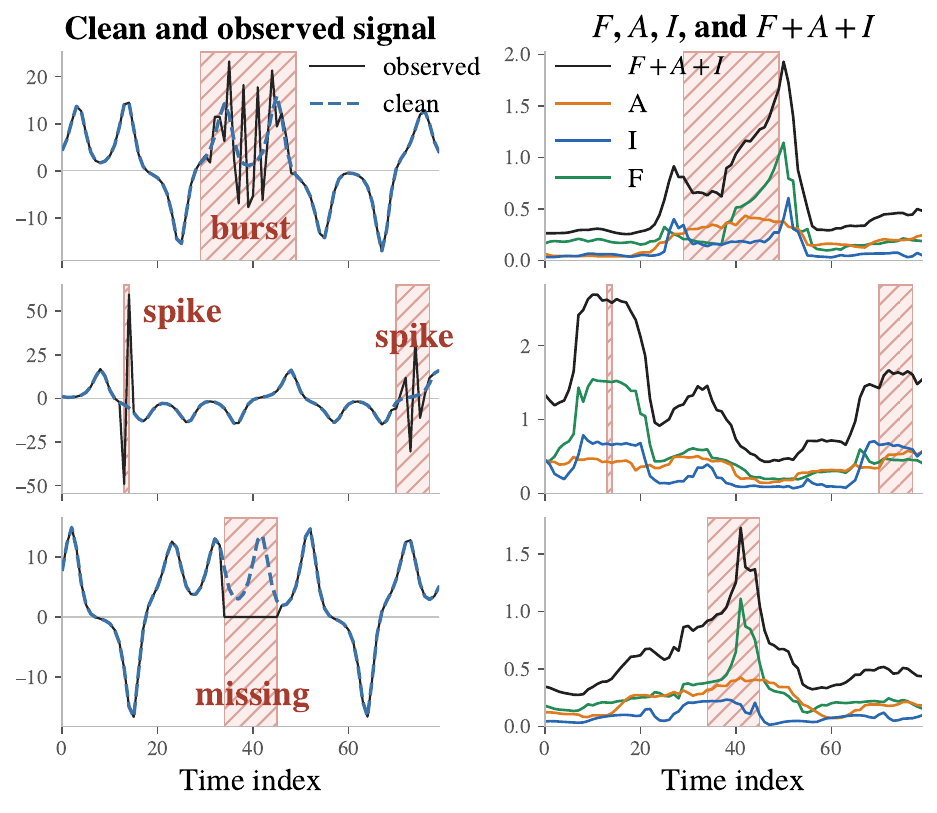}
  \vspace{-15pt}
  \caption{\textbf{Samples from Noisy Lorenz63 (SNR=3).} Clean and observed data with abrupt events are shown alongside analytic features derived from the first IMF of the observed data.}
  \label{fig:lorenz-qai}
  \vspace{-5pt}
\end{wrapfigure}

Table~\ref{tab:exp_settings} summarizes the datasets and the settings. For forecasting, we consider both real-world benchmarks and a synthetic dataset. The synthetic dataset, Noisy Lorenz63, contains evolving underlying dynamics with noise to assess the ability of \method to model time series in which they coexist. We corrupt the input histories but use the corresponding clean Lorenz63 future as the prediction target. Specifically, we use Lorenz63 trajectories \citep{lorenz1963deterministic} as the underlying dynamics and add three noise corruptions: bursts, which represent abruptly emerging high-frequency oscillations; spikes, which represent sharp changes at isolated time points; and missing blocks to simulate periods with no recorded observations, as shown in Figure~\ref{fig:lorenz-qai}. We vary the signal-to-noise ratio (SNR) over $\{3,5,7\}$\,dB to evaluate \method under different noise levels.

\paragraph{Baselines and implementation.}

We integrate \method into six representative SOTA backbones: (1) Transformer-based: Crossformer \citep{crossformer}, iTransformer \citep{itransformer}, PatchTST \citep{patchtst}; (2) MLP-based: TimeMixer \citep{timemixer}; (3) Graph-based: TimeFilter \citep{timefilter}; (4) hybrid architecture: L-Drive \citep{ldrive}, combining CNN, GRU, and MLP modules. To ensure fair evaluation, when applying \method to baseline models, we strictly follow their original hyperparameter settings and only tune the hyperparameters introduced by \method (i.e., $\lambda_1, \lambda_2 \in [0, 1]$, $J \in \{2, 3\}$ and $m_{\max} \in (0,1)$). Table~\ref{tab:main_result_forecasting} uses one fixed seed; two settings are additionally evaluated with three seeds (Appendix~\ref{app:training_seed_stability}). For each classification and anomaly detection experiment, we report means over three independent runs with different seeds. All experiments are conducted using 8 NVIDIA A800 80GB GPUs. See Appendix~\ref{app:experimental_details} for details.

\subsection{Main Results}

\paragraph{Long- and Short-term Forecasting Results.}
\method generally improves all six backbones in both forecasting settings (Table~\ref{tab:main_result_forecasting}). For long-term forecasting, \method reduces TimeMixer's MSE by 14.84\% on Weather, showing that a decomposition-based backbone also benefits from our framework. For short-term forecasting on M4, \method improves all three metrics across all six backbones, reducing Crossformer's MASE and OWA by 2.21\% and 2.02\%, respectively.

\begin{figure}[H]
  \vspace{-15pt}
  \centering
  \includegraphics[width=\linewidth]{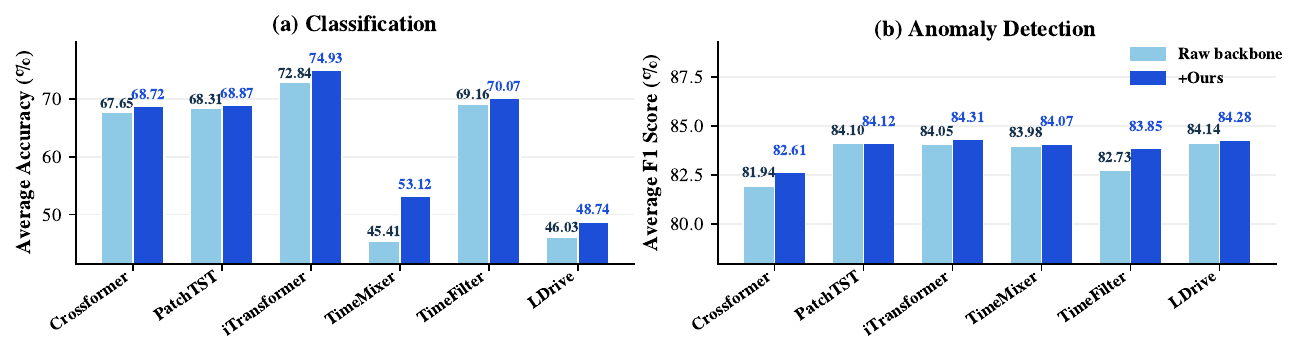}
  \vspace{-15pt}
  \caption{\textbf{Classification and anomaly detection results.} Full results see Table~\ref{tab:full_classification_results} and Table~\ref{tab:full_anomaly_results}.}
  \vspace{-10pt}
  \label{fig:classification_anomaly_averages}
\end{figure}

\paragraph{Classification and anomaly detection results.}
Figure~\ref{fig:classification_anomaly_averages} shows that
\method improves all six backbones on both tasks. For classification, \method raises TimeMixer's accuracy from 45.41\% to 53.12\%, while iTransformer reaches 74.93\%. For anomaly detection, TimeFilter achieves the largest gain in average F1: 1.12 percentage points, from 82.73\% to 83.85\%. These results support the broad applicability of \method across backbone architectures and time series tasks.

\begin{table*}[t]
    \centering
    \vspace{-10pt}
    \caption{\textbf{Long- and short-term forecasting results.} Results are averaged over horizons per dataset, with weighted averages over six M4 subsets. \colorbox{gray!10}{Gray cells} indicate +\method, \textcolor{red}{\textbf{Bold red}} marks the better result within each raw/augmented pair.  Full results are in Table~\ref{tab:full_forecasting_results} and Table~\ref{tab:full_short_results}.}
    \label{tab:main_result_forecasting}
    
    \setlength{\tabcolsep}{1.5pt}
    \renewcommand{\arraystretch}{1}

    \newcommand{\win}[1]{\textcolor{red}{\textbf{#1}}}
    \newcommand{\gc}{\cellcolor{gray!10}}
    \newcommand{\secbest}[1]{#1}
    \newcommand{\imp}[1]{\,($+$#1)}
    \newcommand{\negimp}[1]{\,(#1)}
    
    \resizebox{\textwidth}{!}{%
    \begin{tabular}{@{}c c *{12}{c}@{}}  
    \toprule
    \multirow{2}{*}{\textbf{Dataset}} & \multirow{2}{*}{\textbf{Metric}} & 
    \multicolumn{2}{c}{Crossformer \citeyearpar{crossformer}} & 
    \multicolumn{2}{c}{PatchTST \citeyearpar{patchtst}} & 
    \multicolumn{2}{c}{iTransformer \citeyearpar{itransformer}} & 
    \multicolumn{2}{c}{TimeMixer \citeyearpar{timemixer}} & 
    \multicolumn{2}{c}{TimeFilter \citeyearpar{timefilter}} & 
    \multicolumn{2}{c}{L-Drive \citeyearpar{ldrive}} \\
    \cmidrule(lr){3-4} \cmidrule(lr){5-6} \cmidrule(lr){7-8} \cmidrule(lr){9-10} \cmidrule(lr){11-12} \cmidrule(lr){13-14}
    & & Raw & +\method & Raw & +\method & Raw & +\method & Raw & +\method & Raw & +\method & Raw & +\method \\
    \midrule

    \multirow{2}{*}{ETTh1}
    & MSE
    & \secbest{0.449} & \gc \win{0.443}\imp{1.34\%}
    & \secbest{0.459} & \gc \win{0.457}\imp{0.44\%}
    & \secbest{0.448} & \gc \win{0.440}\imp{1.79\%}
    & \secbest{0.458} & \gc \win{0.453}\imp{1.09\%}
    & \secbest{0.488} & \gc \win{0.475}\imp{2.66\%}
    & \secbest{0.472} & \gc \win{0.465}\imp{1.48\%} \\
    
    & MAE
    & \secbest{0.445} & \gc \win{0.441}\imp{0.90\%}
    & \secbest{0.432} & \gc \win{0.431}\imp{0.23\%}
    & \secbest{0.431} & \gc \win{0.431}\imp{0.00\%}
    & \secbest{0.429} & \gc \win{0.428}\imp{0.23\%}
    & \secbest{0.457} & \gc \win{0.453}\imp{0.88\%}
    & \secbest{0.447} & \gc \win{0.443}\imp{0.89\%} \\
    \addlinespace[0.5em]
    
    \multirow{2}{*}{ETTm1}
    & MSE
    & \secbest{0.413} & \gc \win{0.401}\imp{2.91\%}
    & \secbest{0.396} & \gc \win{0.392}\imp{1.01\%}
    & \secbest{0.400} & \gc \win{0.394}\imp{1.50\%}
    & \secbest{0.393} & \gc \win{0.385}\imp{2.04\%}
    & \win{0.385} & \gc \secbest{0.386}\negimp{-0.26\%}
    & \secbest{0.383} & \gc \win{0.376}\imp{1.83\%} \\
    
    & MAE
    & \secbest{0.404} & \gc \win{0.400}\imp{0.99\%}
    & \secbest{0.387} & \gc \win{0.386}\imp{0.26\%}
    & \secbest{0.390} & \gc \win{0.389}\imp{0.26\%}
    & \secbest{0.385} & \gc \win{0.383}\imp{0.52\%}
    & \win{0.385} & \gc \secbest{0.386}\negimp{-0.26\%}
    & \secbest{0.383} & \gc \win{0.382}\imp{0.26\%} \\
    \addlinespace[0.5em]

    \multirow{2}{*}{Weather}
    & MSE
    & \secbest{0.241} & \gc \win{0.239}\imp{0.83\%}
    & \secbest{0.251} & \gc \win{0.247}\imp{1.59\%}
    & \secbest{0.268} & \gc \win{0.263}\imp{1.87\%}
    & \secbest{0.283} & \gc \win{0.241}\imp{14.84\%}
    & \secbest{0.238} & \gc \win{0.235}\imp{1.26\%}
    & \secbest{0.237} & \gc \win{0.234}\imp{1.27\%} \\
    
    & MAE
    & \secbest{0.268} & \gc \win{0.267}\imp{0.37\%}
    & \secbest{0.270} & \gc \win{0.268}\imp{0.74\%}
    & \secbest{0.280} & \gc \win{0.277}\imp{1.07\%}
    & \secbest{0.302} & \gc \win{0.270}\imp{10.60\%}
    & \secbest{0.261} & \gc \win{0.260}\imp{0.38\%}
    & \secbest{0.258} & \gc \win{0.258}\imp{0.00\%} \\
    \addlinespace[0.5em]

    \multirow{2}{*}{Electricity}
    & MSE
    & \secbest{0.213} & \gc \win{0.205}\imp{3.76\%}
    & \secbest{0.216} & \gc \win{0.213}\imp{1.39\%}
    & \secbest{0.216} & \gc \win{0.212}\imp{1.85\%}
    & \secbest{0.212} & \gc \win{0.202}\imp{4.72\%}
    & \secbest{0.216} & \gc \win{0.208}\imp{3.70\%}
    & \secbest{0.181} & \gc \win{0.179}\imp{1.10\%} \\
    
    & MAE
    & \secbest{0.290} & \gc \win{0.283}\imp{2.41\%}
    & \secbest{0.291} & \gc \win{0.288}\imp{1.03\%}
    & \secbest{0.290} & \gc \win{0.287}\imp{1.03\%}
    & \secbest{0.282} & \gc \win{0.279}\imp{1.06\%}
    & \secbest{0.292} & \gc \win{0.286}\imp{2.05\%}
    & \secbest{0.270} & \gc \win{0.268}\imp{0.74\%} \\
    \addlinespace[0.5em]

    \multirow{2}{*}{ILI}
    & MSE
    & \secbest{5.035} & \gc \win{4.509}\imp{10.45\%}
    & \secbest{3.322} & \gc \win{3.092}\imp{6.92\%}
    & \secbest{3.163} & \gc \win{3.010}\imp{4.84\%}
    & \secbest{3.182} & \gc \win{2.885}\imp{9.33\%}
    & \secbest{2.969} & \gc \win{2.499}\imp{15.83\%}
    & \secbest{3.110} & \gc \win{2.865}\imp{7.88\%} \\
    
    & MAE
    & \secbest{1.542} & \gc \win{1.445}\imp{6.29\%}
    & \secbest{1.110} & \gc \win{1.094}\imp{1.44\%}
    & \secbest{1.069} & \gc \win{1.035}\imp{3.18\%}
    & \secbest{1.149} & \gc \win{1.107}\imp{3.66\%}
    & \secbest{1.033} & \gc \win{0.985}\imp{4.65\%}
    & \secbest{0.969} & \gc \win{0.945}\imp{2.48\%} \\
    \addlinespace[0.5em]

    \multirow{2}{*}{ECG}
    & MSE
    & \secbest{0.771} & \gc \win{0.678}\imp{12.06\%}
    & \secbest{0.577} & \gc \win{0.546}\imp{5.37\%}
    & \secbest{0.654} & \gc \win{0.620}\imp{5.20\%}
    & \secbest{0.633} & \gc \win{0.620}\imp{2.05\%}
    & \secbest{0.579} & \gc \win{0.537}\imp{7.25\%}
    & \secbest{0.664} & \gc \win{0.593}\imp{10.69\%} \\
    
    & MAE
    & \secbest{0.526} & \gc \win{0.506}\imp{3.80\%}
    & \secbest{0.438} & \gc \win{0.411}\imp{6.16\%}
    & \secbest{0.463} & \gc \win{0.456}\imp{1.51\%}
    & \secbest{0.469} & \gc \win{0.455}\imp{2.99\%}
    & \secbest{0.461} & \gc \win{0.416}\imp{9.76\%}
    & \secbest{0.514} & \gc \win{0.462}\imp{10.12\%} \\
    \addlinespace[0.5em]

    \multirow{2}{*}{Laser}
    & MSE
    & \secbest{0.135} & \gc \win{0.117}\imp{13.33\%}
    & \secbest{0.197} & \gc \win{0.173}\imp{12.18\%}
    & \secbest{0.209} & \gc \win{0.182}\imp{12.92\%}
    & \secbest{0.232} & \gc \win{0.201}\imp{13.36\%}
    & \secbest{0.183} & \gc \win{0.149}\imp{18.58\%}
    & \secbest{0.156} & \gc \win{0.148}\imp{5.13\%} \\
    
    & MAE
    & \secbest{0.166} & \gc \win{0.146}\imp{12.05\%}
    & \secbest{0.231} & \gc \win{0.211}\imp{8.66\%}
    & \secbest{0.218} & \gc \win{0.203}\imp{6.88\%}
    & \secbest{0.246} & \gc \win{0.230}\imp{6.50\%}
    & \secbest{0.198} & \gc \win{0.179}\imp{9.60\%}
    & \secbest{0.173} & \gc \win{0.165}\imp{4.62\%} \\
    \addlinespace[0.5em]

    \multirow{2}{*}{\shortstack{Lorenz63\\(SNR=3)}}
    & MSE
    & \secbest{0.690} & \gc \win{0.660}\imp{4.35\%}
    & \secbest{0.850} & \gc \win{0.806}\imp{5.18\%}
    & \secbest{0.815} & \gc \win{0.791}\imp{2.94\%}
    & \secbest{0.837} & \gc \win{0.832}\imp{0.60\%}
    & \secbest{0.700} & \gc \win{0.674}\imp{3.71\%}
    & \secbest{0.726} & \gc \win{0.721}\imp{0.69\%} \\
    
    & MAE
    & \secbest{0.497} & \gc \win{0.469}\imp{5.63\%}
    & \secbest{0.621} & \gc \win{0.589}\imp{5.15\%}
    & \secbest{0.593} & \gc \win{0.575}\imp{3.04\%}
    & \secbest{0.623} & \gc \win{0.620}\imp{0.48\%}
    & \secbest{0.506} & \gc \win{0.490}\imp{3.16\%}
    & \secbest{0.511} & \gc \win{0.510}\imp{0.20\%} \\
    \addlinespace[0.5em]
    
    \multirow{2}{*}{\shortstack{Lorenz63\\(SNR=5)}}
    & MSE
    & \secbest{0.631} & \gc \win{0.581}\imp{7.92\%}
    & \secbest{0.769} & \gc \win{0.752}\imp{2.21\%}
    & \secbest{0.736} & \gc \win{0.720}\imp{2.17\%}
    & \secbest{0.772} & \gc \win{0.763}\imp{1.17\%}
    & \secbest{0.648} & \gc \win{0.595}\imp{8.18\%}
    & \secbest{0.663} & \gc \win{0.656}\imp{1.06\%} \\
    
    & MAE
    & \secbest{0.491} & \gc \win{0.450}\imp{8.35\%}
    & \secbest{0.600} & \gc \win{0.587}\imp{2.17\%}
    & \secbest{0.572} & \gc \win{0.558}\imp{2.45\%}
    & \secbest{0.610} & \gc \win{0.605}\imp{0.82\%}
    & \secbest{0.513} & \gc \win{0.470}\imp{8.38\%}
    & \secbest{0.497} & \gc \win{0.495}\imp{0.40\%} \\
    \addlinespace[0.5em]
    
    \multirow{2}{*}{\shortstack{Lorenz63\\(SNR=7)}}
    & MSE
    & \secbest{0.559} & \gc \win{0.517}\imp{7.51\%}
    & \secbest{0.714} & \gc \win{0.686}\imp{3.92\%}
    & \secbest{0.673} & \gc \win{0.655}\imp{2.67\%}
    & \secbest{0.717} & \gc \win{0.709}\imp{1.12\%}
    & \secbest{0.605} & \gc \win{0.522}\imp{13.72\%}
    & \secbest{0.588} & \gc \win{0.583}\imp{0.85\%} \\
    
    & MAE
    & \secbest{0.462} & \gc \win{0.425}\imp{8.01\%}
    & \secbest{0.585} & \gc \win{0.563}\imp{3.76\%}
    & \secbest{0.549} & \gc \win{0.536}\imp{2.37\%}
    & \secbest{0.591} & \gc \win{0.588}\imp{0.51\%}
    & \secbest{0.509} & \gc \win{0.438}\imp{13.95\%}
    & \secbest{0.474} & \gc \win{0.472}\imp{0.42\%} \\

    % \midrule
    % \multicolumn{2}{c}{\textbf{Avg. Impv.}} &
    % \multicolumn{2}{c}{\textbf{6.33\%} / \textbf{4.53\%}} &
    % \multicolumn{2}{c}{\textbf{4.03\%} / \textbf{2.87\%}} &
    % \multicolumn{2}{c}{\textbf{3.90\%} / \textbf{2.16\%}} &
    % \multicolumn{2}{c}{\textbf{5.47\%} / \textbf{2.98\%}} &
    % \multicolumn{2}{c}{\textbf{7.16\%} / \textbf{4.55\%}} &
    % \multicolumn{2}{c}{\textbf{3.46\%} / \textbf{2.19\%}} \\

    \cmidrule(lr){1-14}
    \multirow{3}{*}{M4}
    & sMAPE
    & \secbest{13.479} & \gc \win{13.231}\imp{1.84\%}
    & \secbest{12.402} & \gc \win{12.349}\imp{0.43\%}
    & \secbest{12.554} & \gc \win{12.513}\imp{0.33\%}
    & \secbest{13.647} & \gc \win{13.413}\imp{1.72\%}
    & \secbest{13.096} & \gc \win{12.845}\imp{1.91\%}
    & \secbest{12.893} & \gc \win{12.716}\imp{1.37\%}
    \\
    
    & MASE
    & \secbest{1.864} & \gc \win{1.822}\imp{2.21\%}
    & \secbest{1.676} & \gc \win{1.662}\imp{0.84\%}
    & \secbest{1.680} & \gc \win{1.676}\imp{0.24\%}
    & \secbest{1.873} & \gc \win{1.828}\imp{2.41\%}
    & \secbest{1.729} & \gc \win{1.695}\imp{1.95\%}
    & \secbest{1.718} & \gc \win{1.682}\imp{2.11\%}
    \\
    
    & OWA
    & \secbest{0.984} & \gc \win{0.964}\imp{2.02\%}
    & \secbest{0.895} & \gc \win{0.890}\imp{0.63\%}
    & \secbest{0.902} & \gc \win{0.900}\imp{0.28\%}
    & \secbest{0.993} & \gc \win{0.973}\imp{2.06\%}
    & \secbest{0.935} & \gc \win{0.917}\imp{1.93\%}
    & \secbest{0.925} & \gc \win{0.909}\imp{1.73\%}
    \\

    \bottomrule
    \end{tabular}%
    }
    \vspace{-1em}
\end{table*}

\subsection{Ablations and Analysis}

\paragraph{Effectiveness of RC.}
\begin{wraptable}{r}{0.57\columnwidth}
\centering
\vspace{-15pt}
\caption{\textbf{Residual correction on Crossformer.}
Noisy Lorenz63 (SNR$=7$ dB, $S=64$); mean(std) over three head seeds.
Trainable/Total: trainable/stored parameters.}
\label{tab:rc_effectiveness}
\newcommand{\win}[1]{\textcolor{red}{\textbf{#1}}}
\newcommand{\secbest}[1]{\underline{#1}}
\resizebox{\linewidth}{!}{%
\begin{tabular}{lcccc}
\toprule
Residual Module
& MSE
& MAE
& Trainable & Total \\
\midrule
No Correction
& 0.815
& 0.655
& --
& -- \\

RC (ours)
& \win{0.805}(0.001)
& \win{0.653}($<$0.001)
& \win{12.48K}
& \win{16.77K} \\

Flatten Linear
& 0.815($<$0.001)
& 0.661($<$0.001)
& 55.49K
& 55.49K \\

Trainable RC
& \secbest{0.809}(0.009)
& \secbest{0.656}(0.006)
& \secbest{16.77K}
& \win{16.77K} \\
\bottomrule
\end{tabular}%
}
\vspace{-15pt}
\end{wraptable} We compare four residual variants: (1) No Correction, (2) RC (ours), (3) Flatten Linear, a direct linear residual mapping, and (4) Trainable RC. For each case, we freeze the same backbone trained with the primary component and train only the residual head. Table~\ref{tab:rc_effectiveness} shows that RC outperforms all alternatives. Flatten Linear's weaker results suggest a potential benefit of modeling temporal dependencies in residual correction. RC does so with only 12.48K trainable parameters.

\paragraph{Decomposition comparison.}
\begin{wraptable}{r}{0.5\columnwidth}
\vspace{-15pt}
\caption{\textbf{Decomposition Comparison.} TimeFilter test errors on noisy Lorenz63. AUC is normalized over the common
energy-reduction range; parentheses give the reduction at minimum error.
See Appendix~\ref{app:decomp_protocol} for the full protocol.}
\label{tab:decomposition_comparison}
\centering
\setlength{\tabcolsep}{2pt}
\renewcommand{\arraystretch}{1.05}
\newcommand{\win}[1]{\textcolor{red}{\textbf{#1}}}
\newcommand{\secbest}[1]{{#1}}

\resizebox{\linewidth}{!}{
\begin{tabular}{c|cc|cc}
\toprule
\multirow{2}{*}{\textbf{Method}} &
\multicolumn{2}{c|}{\textbf{AUC}} &
\multicolumn{2}{c}{\shortstack{\textbf{Minimum Error}}} \\
\cmidrule(lr){2-3}\cmidrule(l){4-5}
& \textbf{MSE} & \textbf{MAE}
& \textbf{MSE} & \textbf{MAE} \\
\midrule
Ours (\method)
& \win{0.340} & \win{0.309}
& \win{0.319} (5.49\%) & \win{0.291} (2.26\%) \\
\midrule
FFT-high
& \secbest{0.352} & \secbest{0.321}
& \secbest{0.322} (1.28\%) & \secbest{0.293} (1.28\%) \\
FFT-low
& 0.403 & 0.358
& 0.347 (1.38\%) & 0.310 (1.38\%) \\
DCT-high
& 0.362 & 0.331
& 0.344 (4.01\%) & 0.310 (4.01\%) \\
STFT-high
& 0.363 & 0.331
& 0.333 (6.30\%) & 0.301 (6.30\%) \\
FFT-adap
& 0.371 & 0.335
& 0.366 (3.30\%) & 0.331 (3.30\%) \\
\bottomrule
\end{tabular}
}

\vspace{-1mm}
\vspace{-6pt}
\end{wraptable}
We isolate decomposition effects using the same TimeFilter backbone and RC, comparing time-adaptive decomposition with FFT high-frequency and low-amplitude truncation (FFT-high/low), DCT and STFT high-frequency truncation (DCT-high, STFT-high), and a sample-adaptive FFT mask (FFT-adap). Table 4 reports test forecasting errors over the shared input-energy reduction range [1.05\%, 9.20\%]. The strongest spectral baseline, FFT-high, performs best at only 1.28\% reduction, suggesting stronger global filtering removes useful patterns. \method remains accurate at higher reductions, suggesting it better preserves forecast-relevant dynamics.

\paragraph{Ablation study.}
We compare raw TimeFilter with a matched raw-input RC control and two analytic-IMF variants, using direct attributes or $\phi$, each with/without RC. Table~\ref{tab:lorenz_ablation} shows that neither analytic processing nor RC alone is uniformly effective: no-RC variants underperform Raw at $\mathrm{SNR}=3$ and $\mathrm{SNR}=5$, while Raw + RC helps only at $\mathrm{SNR}=5$ and $\mathrm{SNR}=7$. Yet RC improves both analytic variants relative to their no-RC counterparts at every SNR, and full \method consistently outperforms Raw + RC. Thus, the results support the complementarity of decomposition and RC and suggest that the gains do not come solely from adding RC to the raw backbone.
\begin{table}[thbp]
\centering
\vspace{-13pt}
\caption{\textbf{Ablation results on Noisy Lorenz63.} Each entry reports MSE / MAE averaged over $S=16, 64$. Analyt. denotes the analytic IMFs. Full results see Appendix~\ref{app:full_ablation}.}
\label{tab:lorenz_ablation}
\resizebox{\linewidth}{!}{%
\begin{tabular}{lcccccc}
      \toprule
      Dataset
      & Raw
      & Raw + RC
      & Analyt. + [Amp., Freq., $O$]
      & Analyt. + $\phi$
      & Analyt. + [Amp., Freq., $O$] + RC
      & \method (Ours) \\
      \midrule
      SNR=3
      & 0.699 / 0.505
      & 0.713 / 0.527
      & 0.713 / 0.522
      & 0.716 / 0.520
      & 0.702 / 0.511
      & \textcolor{red}{\textbf{0.674 / 0.489}} \\
      SNR=5
      & 0.648 / 0.513
      & 0.636 / 0.502
      & 0.663 / 0.522
      & 0.667 / 0.526
      & 0.602 / 0.474
      & \textcolor{red}{\textbf{0.595 / 0.470}} \\
      SNR=7
      & 0.604 / 0.509
      & 0.561 / 0.470
      & 0.600 / 0.500
      & 0.548 / 0.464
      & 0.529 / 0.445
      & \textcolor{red}{\textbf{0.522 / 0.438}} \\
      \bottomrule
    \end{tabular}%
  }
\end{table}
\vspace{-13pt}

\paragraph{Efficiency Analysis.}
Table~\ref{tab:efficiency} compares TimeFilter with and without \method on ILI ($S$=24). \method reduces MSE and MAE by 27.62\% and 7.30\%, respectively, while increasing trainable parameters, peak allocated GPU memory, and model-side FLOPs by only 0.23\%, 0.66\%, and 0.12\%. Under the early-stopping protocol, it requires 22.81\% fewer training iterations. These results show that \method improves forecasting performance with modest additional model-side costs.
\begin{table}[thbp]
\centering
\vspace{-15pt}
\caption{\textbf{Efficiency comparison.} Analytic signals are precomputed offline.}
\label{tab:efficiency}
\setlength{\tabcolsep}{3pt}
\renewcommand{\arraystretch}{1.08}
\newcommand{\win}[1]{#1}
\resizebox{0.9\columnwidth}{!}{%
\begin{tabular}{lcccccc}
\toprule
Method
& Params.
& Peak Alloc. (MiB)
& Train Iters.
& MSE
& MAE
& Model FLOPs/Sample \\
\midrule
Raw
& \win{4,836,628}
& \win{197.24}
& 570
& 3.848
& 1.055
& \win{0.21179} GFLOPs \\
+\method
& 4,847,553
& 198.54
& \win{440}
& \win{2.785}
& \win{0.978}
& 0.21204 GFLOPs \\
\midrule
Change vs. Raw & $\uparrow 0.23\%$ & $\uparrow 0.66\%$
& $\downarrow 22.81\%$ & $\downarrow 27.62\%$
& $\downarrow 7.30\%$ & $\uparrow 0.12\%$ \\
\bottomrule
\end{tabular}%
\vspace{-15pt}
}
\end{table}

\paragraph{Qualitative Analysis.}

\begin{wrapfigure}{r}{0.55\textwidth}
    \centering
    \vspace{-15pt}
    \includegraphics[width=0.55\textwidth]{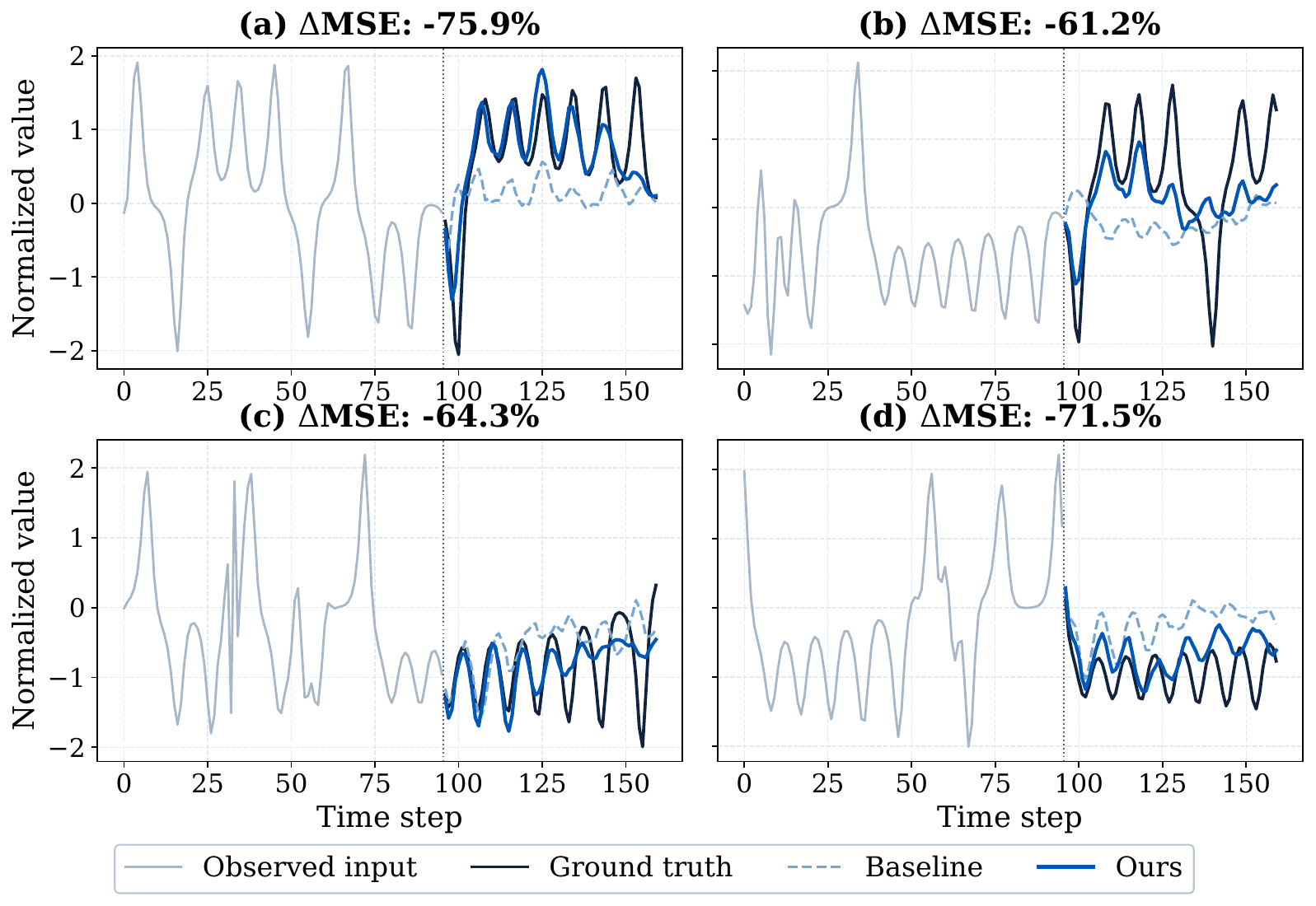}
    \vspace{-15pt}
    \caption{\textbf{Qualitative Visualization.} \method (blue) and backbone (dashed light blue) predictions, with ground truth (dark navy) and observed inputs (gray).}
    \label{fig:lorenz_case_study}
    \vspace{-15pt}
\end{wrapfigure}

As Figure~\ref{fig:lorenz_case_study} shows, under Noisy Lorenz63 (SNR=$7$ dB and $S=64$), the backbone often produces overly smooth trajectories that stay near the mean or show attenuated amplitudes. In contrast, \method effectively exploits the historical context and retains the nonlinear dynamics of the underlying chaotic system. For example, \method better captures the sharp transition at the forecasting boundary and the subsequent recovery in Figure~\ref{fig:lorenz_case_study} (a,b). These results show that \method better captures temporal variation and oscillatory patterns under mixed noise, alleviating the excessive smoothing observed in the backbone.

\paragraph{Decomposition Adaptivity Analysis.}
\begin{wraptable}{r}{0.53\columnwidth}
\vspace{-15pt}
\centering
\scriptsize
\setlength{\tabcolsep}{1.7pt}
\renewcommand{\arraystretch}{1.05}
\caption{\textbf{Decomposition adaptivity.} Temporal-variation alignment is the Pearson correlation between residual energy and local temporal variation.}
\resizebox{\linewidth}{!}{%
\begin{tabular}{@{}lccc@{}}
\toprule
\raisebox{5pt}{\textbf{Dataset}} &
\shortstack{\textbf{Residual}\\\textbf{Magnitude} (\%)} &
\shortstack{\textbf{Residual}\\\textbf{Variance}} &
\shortstack{\textbf{Temporal-Variation}\\\textbf{Alignment}} \\
\midrule
ETTm1 & 1.74 & 0.292 & 0.240 \\
ILI   & 3.70 & 0.370 & 0.484 \\
\bottomrule
\end{tabular}%
}
\label{tab:decomposition_adaptive}
\vspace{-6pt}
\end{wraptable} Using TimeFilter as the backbone, Table~\ref{tab:decomposition_adaptive}
reports results averaged over four prediction lengths for each dataset. Appendix~\ref{app:decomposition_metrics} defines residual magnitude, amplitude variance, and temporal-variation alignment. \method produces residuals on ILI that are larger, more variable, and more strongly correlated with local variation. Together, these results suggest regime-dependent decomposition profiles: residual extraction remains conservative on the regularly periodic ETTm1 series but becomes stronger and more variable on the temporally heterogeneous ILI series.

\paragraph{Hyperparameter Sensitivity.}

\begin{wrapfigure}{r}{0.53\textwidth}
    \centering
    \vspace{-10pt}
    \includegraphics[width=0.53\textwidth]{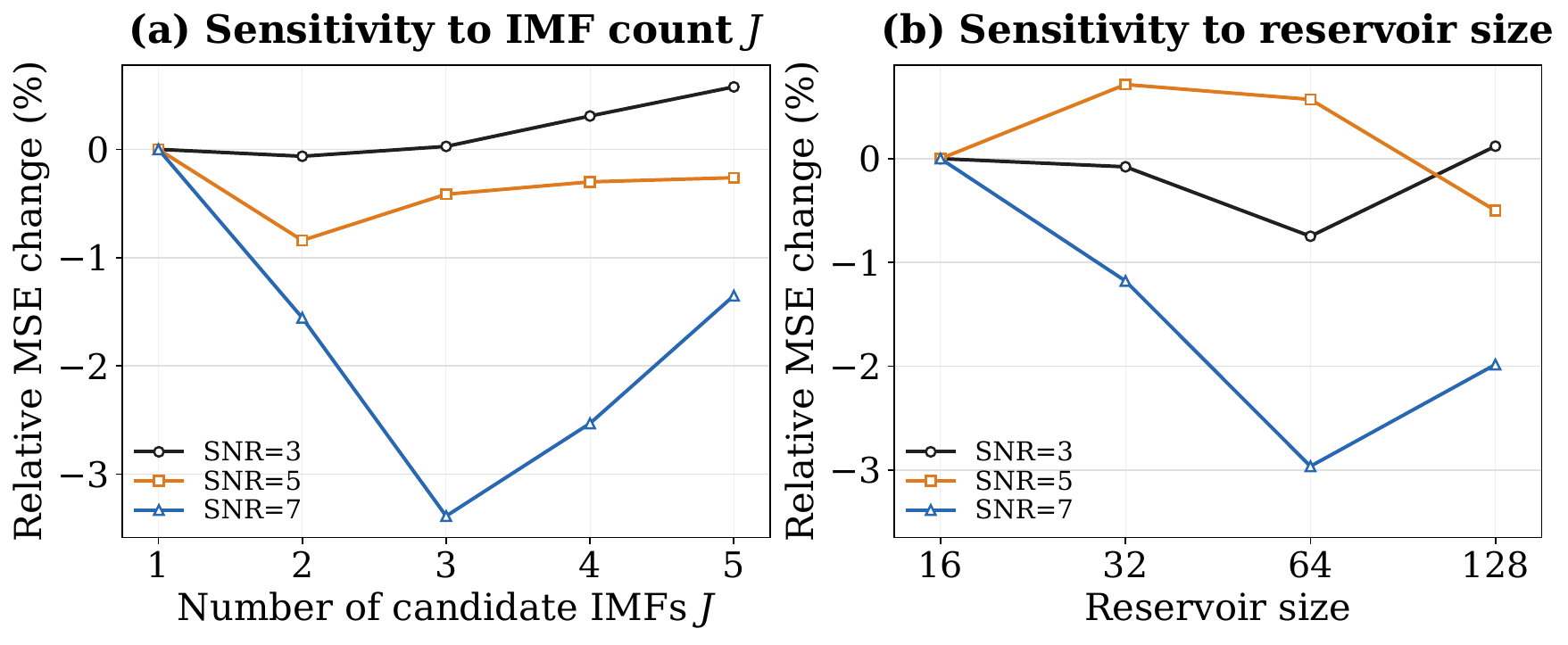}
    \vspace{-15pt}
    \caption{\textbf{Hyperparameter sensitivity.} Relative MSE changes use \(J=1\) in (a) and reservoir size 16 in (b) as references within each SNR.}
    \label{fig:hyperparameter_sensitivity}
    \vspace{-15pt}
\end{wrapfigure}
Figure~\ref{fig:hyperparameter_sensitivity} shows a post-hoc sensitivity analysis of Crossformer (\(S=64\)), with final settings selected on validation data. We omit \(J\geq6\), as many samples lack enough valid IMFs. \(J=1\) and \(J=2\) perform similarly at SNR \(=3\), whereas \(J=2\) and \(J=3\) are optimal at SNR \(=5\) and \(7\), respectively. Larger \(J\) generally raises MSE, consistent with our use of \(J\in\{2,3\}\). For RC, size 64 gives the largest mean relative MSE drop across tested SNRs, though the optimum varies by SNR.

\section{Conclusion and Future Work}
This work addresses the coexistence of evolving dynamics and noise through \textit{time-aware decomposition with correction}. Specifically, \method uses instantaneous-frequency features to extract a primary component that captures evolving dynamics, while residual correction exploits useful dynamics missed by decomposition. Improvements across four time series modeling tasks and six backbones support this approach. Future work will explore online multivariate decomposition to capture evolving cross-variable dynamics without offline preprocessing.

\clearpage

\section*{AI Use Statement}
We used generative AI tools to assist with manuscript writing and polishing, including checks of mathematical notation, dimensions, units, and formula definitions in the main text and appendix. AI tools also helped identify and organize related literature and provided feedback on experimental methodology and the interpretation of results. We manually checked the mathematical expressions and verified cited claims against the original papers. We take responsibility for the final content of this work, including all AI-assisted material.

\bibliography{main}
\bibliographystyle{iclr2027_conference}

\newpage
\appendix
\onecolumn

\section{Experimental Details}
\label{app:experimental_details}

\subsection{Benchmark Details}
\label{app:benchmark_details}

\subsubsection{Long-term Forecasting}

As shown in Table~\ref{tab:long_forecasting_dataset}, we evaluate the performance for long-term forecasting on seven real-world datasets, including ETT datasets (ETTh1, ETTm1), Weather, Electricity, ILI, ECG \citep{ecg,lugovaya2005biometric}, Laser \citep{laser} and a synthetic data Noisy Lorenz63. 

\begin{table*}[thbp]
\centering
\vspace{-10pt}
\caption{Long-term forecasting dataset descriptions. Sizes are reported as (train, validation, test).}
\label{tab:long_forecasting_dataset}
\small
\setlength{\tabcolsep}{2pt}
\renewcommand{\arraystretch}{1.12}
\begin{tabular}{cccccc}
\toprule
Dataset & Dim & Input Length & Prediction Length
& Dataset Size & Frequency / Information \\
\midrule
ETTh1
& 7 & 96 & $\{96,192,336,720\}$
& $(8640,2880,2880)$
& Hourly / Temperature \\

ETTm1
& 7 & 96 & $\{96,192,336,720\}$
& $(34560,11520,11520)$
& 15 min / Temperature \\

Electricity
& 321 & 96 & $\{96,192,336,720\}$
& $(18412,2630,5262)$
& Hourly / Electricity \\

Weather
& 21 & 96 & $\{96,192,336,720\}$
& $(36887,5269,10540)$
& 10 min / Meteorology \\

ILI
& 7 & 24 & $\{24,36,48,60\}$
& $(676,96,194)$
& Weekly / Health \\

ECG
& 1 & 360 & $\{144,288\}$
& $(7000,1000,2000)$
& 500 Hz / Health \\

Laser
& 1 & 100 & $\{50,100\}$
& $(6055,2018,2020)$
& Hourly / Physics \\

\midrule

\makecell[c]{Noisy Lorenz63}
& 3 & 96 & $\{16,64\}$
& $(14000,1999,4001)$
& \makecell[l]{$\Delta t=0.075$ / Chaos} \\
\bottomrule
\end{tabular}
\end{table*}

For the first five public forecasting benchmarks, we follow the standard train, validation, and test splits established in \texttt{BasicTS} \citep{liang2022basicts}. For the remaining datasets, we adopt dedicated forecasting protocols tailored to their signal characteristics, as described below: 
\begin{itemize}[leftmargin=*]
    \item \textbf{ECG}: We use an input length of 360 samples (0.72 s) and prediction lengths of 144 and 288 samples (0.288 s and 0.576 s, respectively). This setting captures sufficient local context to capture cardiac waveform morphology while assessing short-term meaningful horizons.

    \item \textbf{Laser}: We use an input length of 100 to predict the next 50 and 100 observations, evaluating short- and long-horizon forecasting of nonlinear and chaotic dynamics of the laser intensity.

    \item \textbf{Noisy Lorenz63}: We use 96 input steps (a temporal span of 7.2) to predict 16 and 64 steps ahead (horizons of 1.2 and 4.8), evaluating both short-horizon forecasting and long-horizon forecasting under sensitive dependence on initial conditions and mixed noise.
\end{itemize}

All models are trained using Adam \cite{Adam} and a batch size of 64. Training runs for at most 100 epochs, with early stopping after 10 epochs without improvement in validation loss. Within each paired comparison, the raw and +\method variants share the same data splits, backbone configurations, optimization settings, checkpoint-selection rule, and random seed. Given the computational cost of the full benchmark, the main results use one fixed-seed run per configuration. Two selected configurations are additionally evaluated with three seeds, as reported below. Single-run comparisons do not show statistical significance or robustness across seeds; small relative gaps, especially those below $1\%$, warrant caution.

\subsubsection{Short-term Forecasting}
We use M4 for short-term forecasting, with dataset details
provided in Table~\ref{tab:short_forecasting_dataset}.
M4 provides official training and test portions but no separate
validation split. For each series, we reserve the final $H$
observations of its official training portion for validation,
where $H$ is the forecasting horizon.

Following the M4 short-term forecasting protocol
\citep{timesnet,wang2024tssurvey}, all models minimize sMAPE using Adam and a batch size of 32.
Early stopping and checkpoint selection use validation sMAPE,
with an early-stopping patience of 10.
The raw and +\method variants share the same data splits,
backbone configurations, optimization settings, and
checkpoint-selection rule. All additional hyperparameters
are selected on the validation split, and the official
test set is used only for final evaluation.

Each M4 configuration is evaluated in a single run with seed 42.
These matched single-run comparisons do not establish statistical
significance or robustness across seeds. Small relative differences,
particularly those below $1\%$, should therefore be interpreted
cautiously.
\begin{table*}[thbp]
\centering
\vspace{-10pt}
\caption{Short-term forecasting dataset descriptions. Sizes are reported as 
(train, test). M4 does not provide official validation splits; validation subsets are constructed exclusively from their official training
splits as described in the experimental setup.}
\label{tab:short_forecasting_dataset}
\small
\setlength{\tabcolsep}{2pt}
\renewcommand{\arraystretch}{1.12}
\begin{tabular}{cccccc}
\toprule
Dataset & Dim & Input Length & Prediction Length
& Official Series (Train, Test) & Frequency / Information \\
\midrule
M4-Yearly
& 1 & 12 & 6
& $(23{,}000, 23{,}000)$ & Yearly / Demographic \\

M4-Quarterly
& 1 & 16 & 8
& $(24{,}000, 24{,}000)$ & Quarterly / Finance \\

M4-Monthly
& 1 & 36 & 18
& $(48{,}000, 48{,}000)$ & Monthly / Industry \\

M4-Weekly
& 1 & 26 & 13
& $(359, 359)$ & Weekly / Macro \\

M4-Daily
& 1 & 28 & 14
& $(4{,}227, 4{,}227)$ & Daily / Micro \\

M4-Hourly
& 1 & 96 & 48
& $(414, 414)$ & Hourly / Other \\
\bottomrule
\end{tabular}
\vspace{-5pt}
\end{table*}

\subsubsection{Classification and Anomaly Detection}

\paragraph{Classification.}
We evaluate classification on 10 multivariate datasets from
the UEA Time Series Classification Archive \citep{uea},
with details provided in Table~\ref{tab:uea-datasets}.
The archive provides official training and test splits but
no separate validation split. We therefore construct a
deterministic, class-stratified 80/20 training--validation
split from the official training set using seed 42.
This split is fixed across runs, and normalization statistics
are estimated from the resulting training subset only.

PatchTST and iTransformer retain their original classification
implementations. For the other four forecasting backbones
without official classification implementations, we use a
unified masked mean--max classification head.
For each backbone, the raw and \method-augmented variants
use the same head design. In the \method-augmented model,
the backbone processes the primary component to produce
base class logits, while RC maps the residual to a class-wise
logit correction. The correction is added to the base logits
before a single softmax produces the final class probabilities.

All models are trained using Adam,
and a batch size of 64 for at most 100 epochs.
Early stopping and checkpoint selection use validation accuracy,
with an early-stopping patience of 10.
All additional hyperparameters are selected on the validation
split, and the official test set is used only for final
evaluation after model and hyperparameter selection.
Results are averaged over three training seeds: $42$, $43$,
and $44$.

\paragraph{Anomaly detection.}
We evaluate anomaly detection on five datasets:
SMD \citep{smd}, SWaT \citep{swat}, PSM \citep{psm},
and MSL and SMAP \citep{msl_and_smap}.
Table~\ref{tab:anomaly_datasets} provides dataset details.
For each backbone, the baseline and \method-augmented models
share the same backbone configuration and data split.
In the augmented model, the backbone processes the primary
component to produce the base prediction, while RC maps the
residual to a one-step correction. Their outputs are added
to obtain the final prediction.

All models are trained on normal data by minimizing the mean
squared forecasting error. We use Adam and a batch size of 64. Training runs for at most 100 epochs.
Early stopping and checkpoint selection use validation MSE,
with an early-stopping patience of 10.
We repeat training with seeds $42$, $43$, and $44$.

\begin{table}[thbp]
\centering
\caption{UEA classification datasets. Validation sets are split from the official training sets.}
\label{tab:uea-datasets}
\setlength{\tabcolsep}{2.5pt}
\renewcommand{\arraystretch}{1.05}
\begin{tabular}{lccc}
\toprule
\textbf{Dataset} & \textbf{Samples (Train, Test)} & \textbf{Variables} & \textbf{Series Length} \\
\midrule
Handwriting                  & (150, 850)   & 3  & 152 \\
LSST                         & (2459, 2466) & 6  & 36 \\
NATOPS                       & (180, 180)   & 24 & 51 \\
RacketSports                 & (151, 152)   & 6  & 30 \\
SelfRegulationSCP2           & (200, 180)   & 7  & 1152 \\
CharacterTrajectories        & (1422, 1436) & 3  & 60--182 \\
SelfRegulationSCP1           & (268, 293)   & 6  & 896 \\
ArticularyWordRecognition    & (275, 300)   & 9  & 144 \\
JapaneseVowels               & (270, 370)   & 12 & 7--29 \\
UWaveGestureLibrary          & (120, 320)   & 3  & 315 \\
\bottomrule
\end{tabular}
\end{table}

\begin{table}[thbp]
\centering
\caption{Anomaly detection dataset descriptions. Sizes are reported as (train, val, test).}
\label{tab:anomaly_datasets}
\setlength{\tabcolsep}{4pt}
\renewcommand{\arraystretch}{1.0}
\begin{tabular}{cccc}
\hline
\textbf{Dataset} & \textbf{Dataset sizes (train, val, test)} & \textbf{Variable Number} & \textbf{Sliding Window Length} \\
\hline
SMD  & (566724, 141681, 708420) & 38 & 100 \\
MSL  & (44653, 11664, 73729)    & 55 & 100 \\
SMAP & (108146, 27037, 427617)  & 25 & 100 \\
SWaT & (396000, 99000, 449919)  & 51 & 100 \\
PSM  & (105984, 26497, 87841)   & 25 & 100 \\
\hline
\end{tabular}
\end{table}

\subsection{Decomposition Adaptivity Metrics}
\label{app:decomposition_metrics}

We characterize the decomposition output using residual magnitude,
residual-amplitude variance, and temporal-variation alignment. Let
$\mathbf{X}_{h,i},\mathbf{R}'_{h,i}\in\mathbb{R}^{T\times C}$ denote the
normalized input and its decomposition-derived residual for the $i$-th
test window at prediction length $h$, respectively. Here, $T$ and $C$
denote the input length and number of variables, and $N_h$ is the number
of test windows. We use $\epsilon=10^{-12}$ for numerical stability.

\paragraph{Residual magnitude.}
Residual magnitude measures the residual relative to the input:
\begin{equation}
M_h =
\frac{100}{N_h}
\sum_{i=1}^{N_h}
\frac{\left\lVert\mathbf{R}'_{h,i}\right\rVert_F}
     {\left\lVert\mathbf{X}_{h,i}\right\rVert_F+\epsilon}.
\label{eq:residual_magnitude}
\end{equation}
Ignoring the numerical stabilizer, the Frobenius-norm ratio equals the
RMS ratio because the input and residual have the same dimensions. This
measure describes relative residual magnitude rather than an energy
proportion and is therefore not restricted to $[0,100]\%$.

\paragraph{Residual variance.}
We first calculate the residual amplitude at each timestep:
\begin{equation}
a_{h,i,t}
=
\sqrt{
\frac{1}{C}
\sum_{c=1}^{C}
\left(R'_{h,i,t,c}\right)^2
},
\qquad
\bar{a}_{h,i}
=
\frac{1}{T}
\sum_{t=1}^{T}a_{h,i,t}.
\label{eq:residual_amplitude}
\end{equation}
The normalized temporal variance of residual amplitude is defined as
\begin{equation}
V_h =
\frac{1}{N_hT}
\sum_{i=1}^{N_h}
\sum_{t=1}^{T}
\left(
\frac{a_{h,i,t}-\bar{a}_{h,i}}
     {\bar{a}_{h,i}+\epsilon}
\right)^2.
\label{eq:residual_variance}
\end{equation}
Thus, $V_h$ measures how strongly the residual amplitude varies over time
within each input window. A temporally constant residual amplitude,
including an all-zero residual, contributes zero to $V_h$.

\paragraph{Temporal-variation alignment.}
We further measure whether residual extraction is associated with local
input changes. Residual energy and local input variation are defined as
\begin{equation}
e_{h,i,t}
=
\frac{1}{C}
\sum_{c=1}^{C}
\left(R'_{h,i,t,c}\right)^2,
\qquad
d_{h,i,t}
=
\frac{1}{C}
\sum_{c=1}^{C}
\left(
X_{h,i,t,c}-X_{h,i,t-1,c}
\right)^2,
\quad t=2,\ldots,T.
\label{eq:alignment_components}
\end{equation}
Let $\mathcal{I}_h$ denote the set of test windows for which both
$\{e_{h,i,t}\}_{t=2}^{T}$ and $\{d_{h,i,t}\}_{t=2}^{T}$ have nonzero
temporal variance. For $|\mathcal{I}_h|>0$, temporal-variation alignment
is computed as
\begin{equation}
A_h =
\frac{1}{|\mathcal{I}_h|}
\sum_{i\in\mathcal{I}_h}
\operatorname{corr}_{t=2,\ldots,T}
\left(e_{h,i,t},d_{h,i,t}\right),
\qquad |\mathcal{I}_h|>0,
\label{eq:temporal_alignment}
\end{equation}
where $\operatorname{corr}$ denotes the Pearson correlation coefficient.
A larger $A_h$ indicates a more positive average temporal correlation
between residual energy and local input variation. Since local input
changes may contain both valid dynamics and noise, this measure describes
their association rather than noise-identification accuracy. If
$|\mathcal{I}_h|=0$, $A_h$ is treated as undefined rather than zero and
is excluded from the aggregation.

\paragraph{Aggregation.}
Let $\mathcal{H}$ denote the four prediction lengths of a dataset. The
reported residual magnitude and residual-amplitude variance are computed
as
\begin{equation}
\bar{M}
=
\frac{1}{|\mathcal{H}|}
\sum_{h\in\mathcal{H}}M_h,
\qquad
\bar{V}
=
\frac{1}{|\mathcal{H}|}
\sum_{h\in\mathcal{H}}V_h.
\label{eq:metric_aggregation_mv}
\end{equation}
For temporal-variation alignment, let
\begin{equation}
\mathcal{H}_A
=
\left\{
h\in\mathcal{H}:|\mathcal{I}_h|>0
\right\}.
\end{equation}
When $|\mathcal{H}_A|>0$, its dataset-level value is
\begin{equation}
\bar{A}
=
\frac{1}{|\mathcal{H}_A|}
\sum_{h\in\mathcal{H}_A}A_h.
\label{eq:metric_aggregation_alignment}
\end{equation}
If $\mathcal{H}_A$ is empty, $\bar{A}$ is undefined. No empty valid-window
set occurred in the reported experiments.

We use $\mathcal{H}=\{96,192,336,720\}$ for ETTm1 and
$\mathcal{H}=\{24,36,48,60\}$ for ILI. All metrics are first computed
over test windows and then averaged equally across prediction lengths.
All calculations are performed in the normalized model-input space using
TimeFilter as the forecasting backbone.

\subsection{Protocol for the Decomposition Comparison}
\label{app:decomp_protocol}

\paragraph{Training and energy calibration.}
Table~\ref{tab:decomposition_comparison} evaluates TimeFilter on Noisy Lorenz63 (SNR=$7$) with input length 96 and prediction length 16. Each operating point is independently trained using seed 42. All variants share the data split, TimeFilter and RC architectures, loss weights, and optimization protocol. We use Adam, batch size 64, at most 100 epochs, and early stopping with patience 10. For every operating point, the checkpoint with the lowest validation MAE is retained. 

For the validation windows
$\{\mathbf X_i\}_{i=1}^{N_{\rm val}}$ and their estimated removed components
$\{\widehat{\mathbf N}_i\}_{i=1}^{N_{\rm val}}$, we compute
\begin{equation}
r_{\rm val} =
\frac{
\sum_i\|\mathbf X_i\|_F^2-
\sum_i\|\mathbf X_i-\widehat{\mathbf N}_i\|_F^2
}{
\max\!\left(\sum_i\|\mathbf X_i\|_F^2,10^{-12}\right)
}.
\label{eq:net_energy_reduction}
\end{equation}
Thus, $r_{\rm val}$ is a ratio of aggregated energies, rather than the mean
of per-window ratios. The horizontal coordinates in Table~\ref{tab:decomposition_comparison} are measured
on the validation set; the corresponding MSE and MAE are evaluated on the
test set. No test-set energy measurement is used to construct the curves.

\paragraph{Common interval and nAUC.}
For method $m$, let $r_{mj}$ denote its validation-calibrated operating
points. We define the common interval as
\begin{equation}
a=\max\!\left(0.01,\max_m\min_j r_{mj}\right),\qquad
b=\min\!\left(0.10,\min_m\max_j r_{mj}\right),
\end{equation}
which gives $[a,b]=[1.05\%,9.20\%]$. Therefore, every method has measured
points spanning both boundaries. Points immediately outside $[a,b]$ may
bracket a boundary during interpolation, but no extrapolation or endpoint
extension is used.

If multiple configurations produce the same $r_{\rm val}$, their test errors
are averaged; the lowest test error is not selected. We then form a
piecewise-linear test-error curve $\widehat{\mathcal E}_m(r)$. Its normalized
area is computed exactly at the measured knots and interval boundaries:
\begin{equation}
\operatorname{nAUC}_m =
\frac{1}{b-a}
\sum_{k=0}^{K-1}
\frac{\widehat{\mathcal E}_m(s_k)+
      \widehat{\mathcal E}_m(s_{k+1})}{2}
(s_{k+1}-s_k),
\end{equation}
where $\{s_k\}$ contains $a$, $b$, and all measured points within the
interval. Interpolation introduces no additional experimental observations.

The curve minimum is
$\min_{j:r_{mj}\in[a,b]}\mathcal E^{\rm test}_{mj}$ over measured operating
points only. It is reported as an auxiliary descriptive statistic and is
not used for checkpoint or hyperparameter selection. We treat nAUC as the
primary comparison because curve minima remain sensitive to the number of
evaluated configurations.

\begin{table}[t]
\centering
\setlength{\tabcolsep}{3pt}
\caption{Search grids used for Table~4. ``Used'' denotes the number of
operating points within the common energy-reduction interval. Different grid sizes reflect method-specific mappings from control settings to realized energy reduction; nAUC is computed over the same common interval.}
\label{tab:decomp_search}
\begin{tabular}{l l cc}
\toprule
Method & Scanned settings & Runs & Used \\
\midrule
\method
& 11 paired $(m_{\max},m_{\rm init})$ settings
& 11 & 10 \\
FFT-high
& cutoff $\{0.4,0.6\}$; energy target $10^{-4}$--$0.2$
& 10 & 8 \\
FFT-low
& fraction $\{0.4,0.6,0.8\}$; energy target $10^{-4}$--$0.2$
& 11 & 8 \\
DCT-high
& cutoff $\{0.4,0.6\}$; energy target $10^{-4}$--$0.2$
& 10 & 8 \\
STFT-high
& cutoff $\{0.5,0.6\}$; window 24; overlap 0.75
& 9 & 7 \\
FFT-adap
& mask maximum $0.1$--$0.212$; energy target $3{\times}10^{-5}$--$0.02$
& 7 & 4 \\
\bottomrule
\end{tabular}
\end{table}
\section{Complete Results}
\label{app:complete_results}

Due to space constraints, the main text presents aggregated results and selected comparisons. Here we place the full results of all experiments in the following: long-term forecasting in Table~\ref{tab:full_forecasting_results}, short-term forecasting in Table~\ref{tab:full_short_results}, classification in Table~\ref{tab:full_classification_results} and anomaly detection in Table~\ref{tab:full_anomaly_results}.

\subsection{Long-term Forecasting Results}

\paragraph{Results.}
As shown in Table~\ref{tab:full_forecasting_results}, \method consistently improves six structurally different forecasting models on seven real-world datasets, achieving positive gains in 80 out of 84 average comparisons. The improvements are particularly pronounced on ILI, ECG, and Laser, demonstrating its effectiveness for non-stationary, physiological, and chaotic signals. 
Table~\ref{tab:full_forecasting_results} further shows that \method improves all 36 average comparisons on Noisy Lorenz63 across six models and three noise levels. The consistent gains at SNR $=3$ dB demonstrate robustness to severe noise, while the substantial improvements at higher SNRs show that the benefits of \method extend to milder noise conditions.
\begin{table*}[t]
\centering
\newcommand{\win}[1]{\textcolor{red}{\textbf{#1}}}
\newcommand{\secbest}[1]{{#1}}
\caption{
Full long-term forecasting results on seven real-world datasets and Noisy Lorenz63 at three noise levels with and without \method. \colorbox{gray!10}{Gray rows} report the relative improvement ($\Delta$) achieved by \method, where higher values are better.
}
\label{tab:full_forecasting_results}
\resizebox{\textwidth}{!}{
\begin{tabular}{c|c|cccc|cccc|cccc|cccc|cccc|cccc}
\toprule

\multirow{4}{*}{Dataset}
& \multirow{4}{*}{Horizon}
& \multicolumn{4}{c|}{Crossformer}
& \multicolumn{4}{c|}{PatchTST}
& \multicolumn{4}{c|}{iTransformer}
& \multicolumn{4}{c|}{TimeMixer}
& \multicolumn{4}{c|}{TimeFilter}
& \multicolumn{4}{c}{L-Drive}
\\

& & \multicolumn{4}{c|}{\citeyearpar{crossformer}}
& \multicolumn{4}{c|}{\citeyearpar{patchtst}}
& \multicolumn{4}{c|}{\citeyearpar{itransformer}}
& \multicolumn{4}{c|}{\citeyearpar{timemixer}}
& \multicolumn{4}{c|}{\citeyearpar{timefilter}}
& \multicolumn{4}{c}{\citeyearpar{ldrive}}
\\

\cmidrule(lr){3-6}
\cmidrule(lr){7-10}
\cmidrule(lr){11-14}
\cmidrule(lr){15-18}
\cmidrule(lr){19-22}
\cmidrule(lr){23-26}

& & \multicolumn{2}{c}{Raw}
& \multicolumn{2}{c|}{+\method}
& \multicolumn{2}{c}{Raw}
& \multicolumn{2}{c|}{+\method}
& \multicolumn{2}{c}{Raw}
& \multicolumn{2}{c|}{+\method}
& \multicolumn{2}{c}{Raw}
& \multicolumn{2}{c|}{+\method}
& \multicolumn{2}{c}{Raw}
& \multicolumn{2}{c|}{+\method}
& \multicolumn{2}{c}{Raw}
& \multicolumn{2}{c}{+\method}
\\

\cmidrule(lr){3-4}
\cmidrule(lr){5-6}
\cmidrule(lr){7-8}
\cmidrule(lr){9-10}
\cmidrule(lr){11-12}
\cmidrule(lr){13-14}
\cmidrule(lr){15-16}
\cmidrule(lr){17-18}
\cmidrule(lr){19-20}
\cmidrule(lr){21-22}
\cmidrule(lr){23-24}
\cmidrule(lr){25-26}

& & MSE & MAE & MSE & MAE
& MSE & MAE & MSE & MAE
& MSE & MAE & MSE & MAE
& MSE & MAE & MSE & MAE
& MSE & MAE & MSE & MAE
& MSE & MAE & MSE & MAE
\\

\midrule

% ==================== ETTh1 ====================
\multirow{6}{*}{ETTh1}
& 96
& \secbest{0.394} & \secbest{0.404}
& \win{0.383}     & \win{0.397}
& \secbest{0.394} & \win{0.392}
& \win{0.393}     & \win{0.392}
& \secbest{0.384} & \win{0.391}
& \win{0.380}     & \win{0.391}
& \secbest{0.401} & \secbest{0.395}
& \win{0.392}     & \win{0.394}
& \secbest{0.419} & \secbest{0.423}
& \win{0.416}     & \win{0.422}
& \secbest{0.401} & \secbest{0.406}
& \win{0.388}     & \win{0.402}
\\

& 192
& \secbest{0.436} & \secbest{0.431}
& \win{0.424}     & \win{0.423}
& \win{0.447}     & \win{0.423}
& \win{0.447}     & \win{0.423}
& \secbest{0.438} & \secbest{0.422}
& \win{0.431}     & \win{0.420}
& \secbest{0.443} & \win{0.420}
& \win{0.442}     & \win{0.420}
& \secbest{0.482} & \secbest{0.449}
& \win{0.458}     & \win{0.444}
& \secbest{0.449} & \win{0.429}
& \win{0.447}     & \win{0.429}
\\

& 336
& \win{0.471}     & \secbest{0.453}
& \win{0.471}     & \win{0.450}
& \secbest{0.490} & \secbest{0.444}
& \win{0.488}     & \win{0.443}
& \secbest{0.487} & \win{0.446}
& \win{0.475}     & \win{0.446}
& \secbest{0.492} & \secbest{0.441}
& \win{0.484}     & \win{0.440}
& \secbest{0.499} & \win{0.456}
& \win{0.494}     & \win{0.456}
& \secbest{0.499} & \secbest{0.458}
& \win{0.489}     & \win{0.452}
\\

& 720
& \secbest{0.495} & \win{0.493}
& \win{0.494}     & \win{0.493}
& \secbest{0.506} & \secbest{0.470}
& \win{0.501}     & \win{0.466}
& \secbest{0.481} & \secbest{0.466}
& \win{0.473}     & \win{0.465}
& \secbest{0.496} & \secbest{0.460}
& \win{0.493}     & \win{0.458}
& \secbest{0.551} & \secbest{0.500}
& \win{0.531}     & \win{0.491}
& \secbest{0.538} & \secbest{0.493}
& \win{0.536}     & \win{0.488}
\\

\cmidrule(lr){2-26}

& Avg
& 0.449 & 0.445 & 0.443 & 0.441
& 0.459 & 0.432 & 0.457 & 0.431
& 0.448 & 0.431 & 0.440 & 0.431
& 0.458 & 0.429 & 0.453 & 0.428
& 0.488 & 0.457 & 0.475 & 0.453
& 0.472 & 0.447 & 0.465 & 0.443
\\

\rowcolor{gray!20}
\cellcolor{white}
& $\Delta$
& & & 1.34\% & 0.90\%
& & & 0.44\% & 0.23\%
& & & 1.79\% & 0.00\%
& & & 1.09\% & 0.23\%
& & & 2.66\% & 0.88\%
& & & 1.48\% & 0.89\%
\\

\midrule

% ==================== ETTm1 ====================
\multirow{6}{*}{ETTm1}
& 96
& \secbest{0.345} & \secbest{0.362}
& \win{0.330}     & \win{0.356}
& \secbest{0.327} & \win{0.348}
& \win{0.323}     & \win{0.348}
& \secbest{0.330} & \secbest{0.351}
& \win{0.320}     & \win{0.349}
& \secbest{0.318} & \secbest{0.342}
& \win{0.301}     & \win{0.339}
& \win{0.315}     & \secbest{0.346}
& \win{0.315}     & \win{0.344}
& \secbest{0.308} & \secbest{0.341}
& \win{0.304}     & \win{0.340}
\\

& 192
& \secbest{0.378} & \secbest{0.381}
& \win{0.366}     & \win{0.374}
& \secbest{0.374} & \win{0.370}
& \win{0.371}     & \win{0.370}
& \win{0.379}     & \secbest{0.376}
& \win{0.379}     & \win{0.375}
& \secbest{0.373} & \secbest{0.372}
& \win{0.369}     & \win{0.371}
& \win{0.364}     & \win{0.371}
& \secbest{0.365} & \secbest{0.372}
& \secbest{0.361} & \win{0.365}
& \win{0.356}     & \win{0.365}
\\

& 336
& \secbest{0.430} & \secbest{0.411}
& \win{0.418}     & \win{0.407}
& \secbest{0.406} & \secbest{0.395}
& \win{0.403}     & \win{0.394}
& \secbest{0.416} & \secbest{0.399}
& \win{0.407}     & \win{0.398}
& \secbest{0.407} & \secbest{0.394}
& \win{0.402}     & \win{0.393}
& \win{0.397}     & \win{0.392}
& \secbest{0.399} & \secbest{0.395}
& \secbest{0.405} & \secbest{0.396}
& \win{0.390}     & \win{0.393}
\\

& 720
& \secbest{0.498} & \secbest{0.463}
& \win{0.490}     & \win{0.461}
& \secbest{0.477} & \secbest{0.433}
& \win{0.471}     & \win{0.431}
& \secbest{0.473} & \win{0.433}
& \win{0.471}     & \win{0.433}
& \secbest{0.474} & \secbest{0.432}
& \win{0.466}     & \win{0.430}
& \secbest{0.465} & \win{0.432}
& \win{0.464}     & \win{0.432}
& \secbest{0.458} & \secbest{0.431}
& \win{0.452}     & \win{0.430}
\\

\cmidrule(lr){2-26}

& Avg
& 0.413 & 0.404 & 0.401 & 0.400
& 0.396 & 0.387 & 0.392 & 0.386
& 0.400 & 0.390 & 0.394 & 0.389
& 0.393 & 0.385 & 0.385 & 0.383
& 0.385 & 0.385 & 0.386 & 0.386
& 0.383 & 0.383 & 0.376 & 0.382
\\

\rowcolor{gray!20}
\cellcolor{white}
& $\Delta$
& & & 2.91\%  & 0.99\%
& & & 1.01\%  & 0.26\%
& & & 1.50\%  & 0.26\%
& & & 2.04\%  & 0.52\%
& & & -0.26\% & -0.26\%
& & & 1.83\%  & 0.26\%
\\

\midrule

% ==================== Weather ====================
\multirow{6}{*}{Weather}
& 96
& \secbest{0.166} & \secbest{0.204}
& \win{0.161}     & \win{0.202}
& \secbest{0.173} & \secbest{0.208}
& \win{0.168}     & \win{0.205}
& \secbest{0.185} & \secbest{0.217}
& \win{0.177}     & \win{0.211}
& \secbest{0.180} & \secbest{0.226}
& \win{0.162}     & \win{0.210}
& \secbest{0.153} & \secbest{0.191}
& \win{0.150}     & \win{0.190}
& \secbest{0.149} & \win{0.186}
& \win{0.148}     & \win{0.186}
\\

& 192
& \secbest{0.208} & \secbest{0.244}
& \win{0.206}     & \win{0.243}
& \secbest{0.215} & \secbest{0.248}
& \win{0.211}     & \win{0.245}
& \secbest{0.235} & \secbest{0.258}
& \win{0.231}     & \win{0.257}
& \secbest{0.246} & \secbest{0.282}
& \win{0.207}     & \win{0.248}
& \secbest{0.200} & \win{0.237}
& \win{0.199}     & \win{0.237}
& \secbest{0.199} & \secbest{0.234}
& \win{0.197}     & \win{0.233}
\\

& 336
& \secbest{0.260} & \secbest{0.286}
& \win{0.256}     & \win{0.284}
& \secbest{0.269} & \secbest{0.288}
& \win{0.266}     & \win{0.286}
& \secbest{0.291} & \secbest{0.299}
& \win{0.283}     & \win{0.294}
& \secbest{0.309} & \secbest{0.323}
& \win{0.260}     & \win{0.286}
& \secbest{0.261} & \secbest{0.282}
& \win{0.257}     & \win{0.281}
& \secbest{0.259} & \secbest{0.279}
& \win{0.256}     & \win{0.278}
\\

& 720
& \win{0.331}     & \secbest{0.339}
& \win{0.331}     & \win{0.337}
& \secbest{0.345} & \secbest{0.337}
& \win{0.344}     & \win{0.336}
& \secbest{0.362} & \secbest{0.345}
& \win{0.359}     & \win{0.344}
& \secbest{0.395} & \secbest{0.375}
& \win{0.336}     & \win{0.334}
& \secbest{0.337} & \win{0.332}
& \win{0.335}     & \secbest{0.333}
& \secbest{0.339} & \win{0.333}
& \win{0.334}     & \win{0.333}
\\

\cmidrule(lr){2-26}

& Avg
& 0.241 & 0.268 & 0.239 & 0.267
& 0.251 & 0.270 & 0.247 & 0.268
& 0.268 & 0.280 & 0.263 & 0.277
& 0.283 & 0.302 & 0.241 & 0.270
& 0.238 & 0.261 & 0.235 & 0.260
& 0.237 & 0.258 & 0.234 & 0.258
\\

\rowcolor{gray!20}
\cellcolor{white}
& $\Delta$
& & & 0.83\%  & 0.37\%
& & & 1.59\%  & 0.74\%
& & & 1.87\%  & 1.07\%
& & & 14.84\% & 10.60\%
& & & 1.26\%  & 0.38\%
& & & 1.27\%  & 0.00\%
\\

\midrule

% ==================== Electricity ====================
\multirow{6}{*}{Electricity}
& 96
& \secbest{0.192} & \secbest{0.267}
& \win{0.178}     & \win{0.257}
& \secbest{0.194} & \secbest{0.268}
& \win{0.191}     & \win{0.266}
& \secbest{0.193} & \secbest{0.267}
& \win{0.190}     & \win{0.263}
& \secbest{0.195} & \secbest{0.263}
& \win{0.181}     & \win{0.259}
& \secbest{0.193} & \secbest{0.268}
& \win{0.187}     & \win{0.263}
& \secbest{0.156} & \win{0.246}
& \win{0.152}     & \win{0.246}
\\

& 192
& \secbest{0.198} & \win{0.275}
& \win{0.196}     & \win{0.275}
& \secbest{0.199} & \secbest{0.276}
& \win{0.195}     & \win{0.272}
& \secbest{0.198} & \secbest{0.275}
& \win{0.192}     & \win{0.272}
& \secbest{0.193} & \secbest{0.267}
& \win{0.185}     & \win{0.264}
& \secbest{0.199} & \secbest{0.276}
& \win{0.193}     & \win{0.271}
& \secbest{0.170} & \win{0.259}
& \win{0.169}     & \win{0.259}
\\

& 336
& \secbest{0.214} & \secbest{0.295}
& \win{0.208}     & \win{0.288}
& \secbest{0.214} & \secbest{0.292}
& \win{0.213}     & \win{0.290}
& \secbest{0.214} & \secbest{0.292}
& \win{0.211}     & \win{0.289}
& \secbest{0.209} & \secbest{0.283}
& \win{0.201}     & \win{0.280}
& \secbest{0.213} & \secbest{0.293}
& \win{0.201}     & \win{0.287}
& \secbest{0.182} & \win{0.270}
& \win{0.181}     & \win{0.270}
\\

& 720
& \secbest{0.249} & \secbest{0.324}
& \win{0.237}     & \win{0.312}
& \secbest{0.255} & \secbest{0.326}
& \win{0.252}     & \win{0.322}
& \secbest{0.257} & \secbest{0.327}
& \win{0.253}     & \win{0.323}
& \secbest{0.249} & \secbest{0.315}
& \win{0.239}     & \win{0.311}
& \secbest{0.260} & \secbest{0.331}
& \win{0.252}     & \win{0.323}
& \secbest{0.215} & \secbest{0.303}
& \win{0.213}     & \win{0.296}
\\

\cmidrule(lr){2-26}

& Avg
& 0.213 & 0.290 & 0.205 & 0.283
& 0.216 & 0.291 & 0.213 & 0.288
& 0.216 & 0.290 & 0.212 & 0.287
& 0.212 & 0.282 & 0.202 & 0.279
& 0.216 & 0.292 & 0.208 & 0.286
& 0.181 & 0.270 & 0.179 & 0.268
\\

\rowcolor{gray!20}
\cellcolor{white}
& $\Delta$
& & & 3.76\% & 2.41\%
& & & 1.39\% & 1.03\%
& & & 1.85\% & 1.03\%
& & & 4.72\% & 1.06\%
& & & 3.70\% & 2.05\%
& & & 1.10\% & 0.74\%
\\

\midrule

% === ILI ===
\multirow{6}{*}{ILI}
& 24
& \secbest{4.736} & \secbest{1.480} & \win{4.536} & \win{1.417}
& \secbest{3.633} & \secbest{1.079} & \win{3.126} & \win{1.061}
& \secbest{3.507} & \secbest{1.071} & \win{3.231} & \win{1.023}
& \secbest{3.124} & \secbest{1.136} & \win{2.758} & \win{1.079}
& \secbest{3.848} & \secbest{1.055} & \win{2.785} & \win{0.978}
& \secbest{3.848} & \secbest{0.940} & \win{3.290} & \win{0.920} \\

& 36
& \secbest{5.153} & \secbest{1.561} & \win{4.403} & \win{1.423}
& \secbest{4.019} & \secbest{1.192} & \win{3.824} & \win{1.159}
& \secbest{3.974} & \secbest{1.152} & \win{3.935} & \win{1.146}
& \secbest{3.538} & \secbest{1.214} & \win{3.134} & \win{1.144}
& \secbest{2.916} & \secbest{1.032} & \win{2.735} & \win{1.007}
& \secbest{4.324} & \secbest{1.095} & \win{4.225} & \win{1.070} \\

& 48
& \secbest{5.244} & \secbest{1.576} & \win{4.267} & \win{1.409}
& \secbest{2.939} & \secbest{1.099} & \win{2.812} & \win{1.086}
& \secbest{2.513} & \secbest{1.005} & \win{2.439} & \win{0.976}
& \secbest{3.055} & \secbest{1.130} & \win{2.806} & \win{1.097}
& \secbest{2.744} & \secbest{1.037} & \win{2.487} & \win{1.020}
& \secbest{2.241} & \secbest{0.926} & \win{2.149} & \win{0.917} \\

& 60
& \secbest{5.006} & \secbest{1.550} & \win{4.831} & \win{1.529}
& \secbest{2.695} & \win{1.071} & \win{2.607} & \win{1.071}
& \secbest{2.657} & \secbest{1.049} & \win{2.435} & \win{0.994}
& \secbest{3.010} & \secbest{1.114} & \win{2.843} & \win{1.108}
& \secbest{2.369} & \secbest{1.007} & \win{1.988} & \win{0.935}
& \secbest{2.025} & \secbest{0.916} & \win{1.794} & \win{0.871} \\

\cmidrule(lr){2-26}

& Avg
& 5.035 & 1.542 & 4.509 & 1.445
& 3.322 & 1.110 & 3.092 & 1.094
& 3.163 & 1.069 & 3.010 & 1.035
& 3.182 & 1.149 & 2.885 & 1.107
& 2.969 & 1.033 & 2.499 & 0.985
& 3.110 & 0.969 & 2.865 & 0.945 \\

\rowcolor{gray!20}
\cellcolor{white}
& $\Delta$
& & & 10.45\% & 6.29\%
& & & 6.92\% & 1.44\%
& & & 4.84\% & 3.18\%
& & & 9.33\% & 3.66\%
& & & 15.83\% & 4.65\%
& & & 7.88\% & 2.48\% \\

\midrule

% === ECG ===
\multirow{4}{*}{ECG}
& 144
& \secbest{0.699} & \secbest{0.473} & \win{0.572} & \win{0.466}
& \secbest{0.540} & \secbest{0.377} & \win{0.513} & \win{0.365}
& \secbest{0.630} & \secbest{0.406} & \win{0.579} & \win{0.397}
& \secbest{0.587} & \secbest{0.396} & \win{0.581} & \win{0.383}
& \secbest{0.551} & \secbest{0.449} & \win{0.489} & \win{0.387}
& \secbest{0.582} & \secbest{0.424} & \win{0.540} & \win{0.404} \\

& 288
& \secbest{0.843} & \secbest{0.579} & \win{0.783} & \win{0.546}
& \secbest{0.613} & \secbest{0.498} & \win{0.579} & \win{0.457}
& \secbest{0.677} & \secbest{0.519} & \win{0.661} & \win{0.515}
& \secbest{0.679} & \secbest{0.542} & \win{0.659} & \win{0.527}
& \secbest{0.607} & \secbest{0.472} & \win{0.584} & \win{0.444}
& \secbest{0.746} & \secbest{0.604} & \win{0.645} & \win{0.520} \\

\cmidrule(lr){2-26}

& Avg
& 0.771 & 0.526 & 0.678 & 0.506
& 0.577 & 0.438 & 0.546 & 0.411
& 0.654 & 0.463 & 0.620 & 0.456
& 0.633 & 0.469 & 0.620 & 0.455
& 0.579 & 0.461 & 0.537 & 0.416
& 0.664 & 0.514 & 0.593 & 0.462 \\

\rowcolor{gray!20}
\cellcolor{white}
& $\Delta$
& & & 12.06\% & 3.80\%
& & & 5.37\% & 6.16\%
& & & 5.20\% & 1.51\%
& & & 2.05\% & 2.99\%
& & & 7.25\% & 9.76\%
& & & 10.69\% & 10.12\% \\

\midrule

% === Laser ===
\multirow{4}{*}{Laser}
& 50
& \secbest{0.106} & \secbest{0.147} & \win{0.072} & \win{0.114}
& \secbest{0.140} & \secbest{0.182} & \win{0.124} & \win{0.166}
& \secbest{0.150} & \secbest{0.166} & \win{0.120} & \win{0.147}
& \secbest{0.170} & \secbest{0.194} & \win{0.144} & \win{0.177}
& \secbest{0.126} & \secbest{0.152} & \win{0.088} & \win{0.129}
& \secbest{0.106} & \secbest{0.132} & \win{0.092} & \win{0.118} \\

& 100
& \secbest{0.163} & \secbest{0.185} & \win{0.162} & \win{0.178}
& \secbest{0.254} & \secbest{0.280} & \win{0.221} & \win{0.255}
& \secbest{0.267} & \secbest{0.270} & \win{0.244} & \win{0.258}
& \secbest{0.293} & \secbest{0.298} & \win{0.257} & \win{0.282}
& \secbest{0.239} & \secbest{0.244} & \win{0.210} & \win{0.229}
& \secbest{0.205} & \secbest{0.214} & \win{0.204} & \win{0.212} \\

\cmidrule(lr){2-26}

& Avg
& 0.135 & 0.166 & 0.117 & 0.146
& 0.197 & 0.231 & 0.173 & 0.211
& 0.209 & 0.218 & 0.182 & 0.203
& 0.232 & 0.246 & 0.201 & 0.230
& 0.183 & 0.198 & 0.149 & 0.179
& 0.156 & 0.173 & 0.148 & 0.165 \\

\rowcolor{gray!20}
\cellcolor{white}
& $\Delta$
& & & 13.33\% & 12.05\%
& & & 12.18\% & 8.66\%
& & & 12.92\% & 6.88\%
& & & 13.36\% & 6.50\%
& & & 18.58\% & 9.60\%
& & & 5.13\% & 4.62\% \\

\midrule

% ==================== Lorenz63 (SNR=3) ====================
\multirow{4}{*}{\shortstack{Lorenz63\\(SNR=3)}}
& 16
& \secbest{0.490} & \secbest{0.333}
& \win{0.483}     & \win{0.327}
& \secbest{0.718} & \secbest{0.527}
& \win{0.674}     & \win{0.493}
& \secbest{0.672} & \secbest{0.489}
& \win{0.644}     & \win{0.469}
& \secbest{0.742} & \secbest{0.556}
& \win{0.732}     & \win{0.550}
& \secbest{0.551} & \secbest{0.391}
& \win{0.508}     & \win{0.363}
& \secbest{0.532} & \secbest{0.364}
& \win{0.527}     & \win{0.363}
\\

& 64
& \secbest{0.889} & \secbest{0.661}
& \win{0.836}     & \win{0.610}
& \secbest{0.981} & \secbest{0.715}
& \win{0.937}     & \win{0.685}
& \secbest{0.958} & \secbest{0.697}
& \win{0.937}     & \win{0.680}
& \win{0.931}     & \win{0.689}
& \win{0.931}     & \win{0.689}
& \secbest{0.848} & \secbest{0.620}
& \win{0.839}     & \win{0.616}
& \secbest{0.920} & \secbest{0.657}
& \win{0.914}     & \win{0.656}
\\

\cmidrule(lr){2-26}

& Avg
& 0.690 & 0.497 & 0.660 & 0.469
& 0.850 & 0.621 & 0.806 & 0.589
& 0.815 & 0.593 & 0.791 & 0.575
& 0.837 & 0.623 & 0.832 & 0.620
& 0.700 & 0.506 & 0.674 & 0.490
& 0.726 & 0.511 & 0.721 & 0.510
\\

\rowcolor{gray!20}
\cellcolor{white}
& $\Delta$
& & & 4.35\% & 5.63\%
& & & 5.18\% & 5.15\%
& & & 2.94\% & 3.04\%
& & & 0.60\% & 0.48\%
& & & 3.71\% & 3.16\%
& & & 0.69\% & 0.20\%
\\

\midrule

% ==================== Lorenz63 (SNR=5) ====================
\multirow{4}{*}{\shortstack{Lorenz63\\(SNR=5)}}
& 16
& \secbest{0.410} & \secbest{0.320}
& \win{0.394}     & \win{0.304}
& \secbest{0.626} & \secbest{0.500}
& \win{0.612}     & \win{0.490}
& \secbest{0.574} & \secbest{0.459}
& \win{0.555}     & \win{0.442}
& \secbest{0.654} & \secbest{0.531}
& \win{0.640}     & \win{0.524}
& \secbest{0.508} & \secbest{0.415}
& \win{0.411}     & \win{0.332}
& \secbest{0.444} & \win{0.341}
& \win{0.439}     & \win{0.341}
\\

& 64
& \secbest{0.852} & \secbest{0.662}
& \win{0.768}     & \win{0.595}
& \secbest{0.912} & \secbest{0.700}
& \win{0.891}     & \win{0.683}
& \secbest{0.897} & \secbest{0.684}
& \win{0.885}     & \win{0.674}
& \secbest{0.890} & \secbest{0.688}
& \win{0.886}     & \win{0.685}
& \secbest{0.788} & \secbest{0.611}
& \win{0.778}     & \win{0.608}
& \secbest{0.881} & \secbest{0.652}
& \win{0.873}     & \win{0.649}
\\

\cmidrule(lr){2-26}

& Avg
& 0.631 & 0.491 & 0.581 & 0.450
& 0.769 & 0.600 & 0.752 & 0.587
& 0.736 & 0.572 & 0.720 & 0.558
& 0.772 & 0.610 & 0.763 & 0.605
& 0.648 & 0.513 & 0.595 & 0.470
& 0.663 & 0.497 & 0.656 & 0.495
\\

\rowcolor{gray!20}
\cellcolor{white}
& $\Delta$
& & & 7.92\% & 8.35\%
& & & 2.21\% & 2.17\%
& & & 2.17\% & 2.45\%
& & & 1.17\% & 0.82\%
& & & 8.18\% & 8.38\%
& & & 1.06\% & 0.40\%
\\

\midrule

% ==================== Lorenz63 (SNR=7) ====================
\multirow{4}{*}{\shortstack{Lorenz63\\(SNR=7)}}
& 16
& \secbest{0.303} & \secbest{0.268}
& \win{0.293}     & \win{0.252}
& \secbest{0.560} & \secbest{0.481}
& \win{0.524}     & \win{0.453}
& \secbest{0.496} & \secbest{0.429}
& \win{0.469}     & \win{0.410}
& \secbest{0.581} & \secbest{0.504}
& \win{0.568}     & \win{0.499}
& \secbest{0.414} & \secbest{0.377}
& \win{0.320}     & \win{0.289}
& \secbest{0.362} & \secbest{0.312}
& \win{0.357}     & \win{0.309}
\\

& 64
& \secbest{0.815} & \secbest{0.655}
& \win{0.741}     & \win{0.598}
& \secbest{0.868} & \secbest{0.688}
& \win{0.847}     & \win{0.672}
& \secbest{0.849} & \secbest{0.669}
& \win{0.841}     & \win{0.662}
& \secbest{0.852} & \secbest{0.678}
& \win{0.850}     & \win{0.677}
& \secbest{0.795} & \secbest{0.640}
& \win{0.724}     & \win{0.586}
& \secbest{0.814} & \secbest{0.635}
& \win{0.808}     & \win{0.634}
\\

\cmidrule(lr){2-26}

& Avg
& 0.559 & 0.462 & 0.517 & 0.425
& 0.714 & 0.585 & 0.686 & 0.563
& 0.673 & 0.549 & 0.655 & 0.536
& 0.717 & 0.591 & 0.709 & 0.588
& 0.605 & 0.509 & 0.522 & 0.438
& 0.588 & 0.474 & 0.583 & 0.472
\\

\rowcolor{gray!20}
\cellcolor{white}
& $\Delta$
& & & 7.51\%  & 8.01\%
& & & 3.92\%  & 3.76\%
& & & 2.67\%  & 2.37\%
& & & 1.12\%  & 0.51\%
& & & 13.72\% & 13.95\%
& & & 0.85\%  & 0.42\%
\\
\bottomrule
\end{tabular}
}
\end{table*}

\paragraph{Training-seed stability.}
\label{app:training_seed_stability}
We further evaluate training-seed variability on two representative
forecasting settings: TimeFilter on ILI with a prediction horizon of 60,
representing a relatively large observed improvement, and PatchTST on ETTh1
with a prediction horizon of 96, representing a typical small-gain setting.
We perform full end-to-end training using seeds
$\{42,43,44\}$. Within each seed, the Raw and $+\method$ variants use
identical data splits, backbone configurations, optimization settings,
validation-MAE checkpoint-selection rules, and physical GPUs.

On TimeFilter--ILI, $+\method$ changes MSE from $2.387\pm0.402$ to $2.000\pm0.173$ and MAE from $1.002\pm0.074$ to $0.934\pm0.035$, corresponding to mean paired reductions of $15.01\%\pm12.73\%$ and $6.43\%\pm6.54\%$, respectively.
MSE improves in all three runs, while MAE improves in two of the three runs.
On PatchTST--ETTh1, $+\method$ changes MSE from $0.3995\pm0.0101$ to $0.3985\pm0.0098$ and MAE from $0.3957\pm0.0045$ to $0.3956\pm0.0043$, corresponding to mean paired changes of $0.26\%\pm0.09\%$ and $0.02\%\pm0.05\%$, respectively. Across the three seeds, \method consistently reduces MSE in both settings, with particularly substantial gains on TimeFilter--ILI. The average
MAE is also improved, indicating that the benefits extend across both
evaluation metrics. All deviations denote sample standard deviations
across the three runs.

\subsection{Short-term Forecasting Results}
\paragraph{Setups.}
Performance is evaluated using sMAPE, MASE, and OWA, where lower values indicate better forecasting accuracy. OWA measures performance relative to the official Naive2 baseline. Overall M4 results are aggregated across the six frequency subsets, weighted by the number of series in each subset.

\paragraph{Results.}
\begin{table*}[thbp]
\centering
\newcommand{\gc}{\cellcolor{gray!10}}
\newcommand{\win}[1]{\textcolor{red}{\textbf{#1}}}
\newcommand{\secbest}[1]{#1}
\newcommand{\impv}[2]{%
  \shortstack[c]{%
    #1\\[-1pt]
    {\scriptsize\textcolor{black!65}{(+#2)}}%
  }%
}
\newcommand{\imp}[1]{\,(\ensuremath{+}#1)}
\newcommand{\modelyear}[2]{\shortstack{#1\\[-2pt]\citeyearpar{#2}}}
\newcommand{\negimp}[1]{\,(#1)}
\caption{
Full short-term forecasting results on M4 with and without \method. \colorbox{gray!10}{Gray cells} report the results obtained with \method and the corresponding relative improvement.}
\label{tab:full_short_results}
\resizebox{\textwidth}{!}{
\begin{tabular}{c|c|cc|cc|cc|cc|cc|cc}
\toprule
\multirow{2}{*}{\textbf{Subset}}
& \multirow{2}{*}{\textbf{Metric}}
& \multicolumn{2}{c|}{\modelyear{Crossformer}{crossformer}}
& \multicolumn{2}{c|}{\modelyear{PatchTST}{patchtst}}
& \multicolumn{2}{c|}{\modelyear{iTransformer}{itransformer}}
& \multicolumn{2}{c|}{\modelyear{TimeMixer}{timemixer}}
& \multicolumn{2}{c|}{\modelyear{TimeFilter}{timefilter}}
& \multicolumn{2}{c}{\modelyear{L-Drive}{ldrive}}
\\

\cmidrule(lr){3-4}
\cmidrule(lr){5-6}
\cmidrule(lr){7-8}
\cmidrule(lr){9-10}
\cmidrule(lr){11-12}
\cmidrule(lr){13-14}

& & Raw & $+$\method
  & Raw & $+$\method
  & Raw & $+$\method
  & Raw & $+$\method
  & Raw & $+$\method
  & Raw & $+$\method
\\
\midrule
% ==================== Yearly ====================
\multirow{3}{*}{Yearly}
& sMAPE
& 15.369 & \gc 15.059
& 13.926 & \gc 13.859
& 14.076 & \gc 14.023
& 15.343 & \gc 15.169
& 13.996 & \gc 14.032
& 14.076 & \gc 13.896
\\
& MASE
& 3.571 & \gc 3.442
& 3.177 & \gc 3.125
& 3.169 & \gc 3.159
& 3.450 & \gc 3.368
& 3.115 & \gc 3.111
& 3.177 & \gc 3.102
\\
& OWA
& 0.919 & \gc 0.894
& 0.826 & \gc 0.817
& 0.829 & \gc 0.827
& 0.904 & \gc 0.888
& 0.820 & \gc 0.821
& 0.830 & \gc 0.815
\\
\midrule

% ==================== Quarterly ====================
\multirow{3}{*}{Quarterly}
& sMAPE
& 11.467 & \gc 11.588
& 10.640 & \gc 10.664
& 10.834 & \gc 10.741
& 12.686 & \gc 12.048
& 11.289 & \gc 10.971
& 11.151 & \gc 10.899
\\
& MASE
& 1.302 & \gc 1.369
& 1.257 & \gc 1.257
& 1.260 & \gc 1.251
& 1.552 & \gc 1.460
& 1.326 & \gc 1.264
& 1.371 & \gc 1.306
\\
& OWA
& 0.995 & \gc 1.025
& 0.942 & \gc 0.942
& 0.951 & \gc 0.944
& 1.142 & \gc 1.079
& 0.996 & \gc 0.959
& 1.006 & \gc 0.971
\\
\midrule

% ==================== Monthly ====================
\multirow{3}{*}{Monthly}
& sMAPE
& 14.388 & \gc 14.011
& 13.329 & \gc 13.241
& 13.493 & \gc 13.478
& 14.200 & \gc 14.127
& 14.397 & \gc 14.014
& 14.043 & \gc 13.883
\\
& MASE
& 1.102 & \gc 1.086
& 1.000 & \gc 0.993
& 1.019 & \gc 1.017
& 1.109 & \gc 1.106
& 1.087 & \gc 1.046
& 1.038 & \gc 1.026
\\
& OWA
& 1.017 & \gc 0.996
& 0.932 & \gc 0.926
& 0.947 & \gc 0.945
& 1.013 & \gc 1.010
& 1.010 & \gc 0.978
& 0.975 & \gc 0.964
\\
\midrule

% ==================== Weekly ====================
\multirow{3}{*}{Weekly}
& sMAPE
& 10.985 & \gc 10.177
& 9.030 & \gc 8.964
& 9.551 & \gc 9.474
& 10.324 & \gc 10.302
& 9.594 & \gc 9.875
& 9.727 & \gc 9.869
\\
& MASE
& 3.075 & \gc 3.254
& 2.544 & \gc 2.539
& 2.633 & \gc 2.645
& 3.302 & \gc 3.337
& 2.907 & \gc 2.793
& 2.897 & \gc 2.821
\\
& OWA
& 1.153 & \gc 1.141
& 0.951 & \gc 0.946
& 0.995 & \gc 0.993
& 1.158 & \gc 1.163
& 1.047 & \gc 1.042
& 1.052 & \gc 1.046
\\
\midrule

% ==================== Daily ====================
\multirow{3}{*}{Daily}
& sMAPE
& 3.732 & \gc 3.344
& 3.233 & \gc 3.246
& 3.172 & \gc 3.191
& 3.359 & \gc 3.322
& 3.340 & \gc 3.400
& 3.137 & \gc 3.181
\\
& MASE
& 4.136 & \gc 3.662
& 3.416 & \gc 3.473
& 3.386 & \gc 3.403
& 3.630 & \gc 3.589
& 3.598 & \gc 3.672
& 3.363 & \gc 3.425
\\
& OWA
& 1.244 & \gc 1.108
& 1.052 & \gc 1.063
& 1.037 & \gc 1.043
& 1.105 & \gc 1.093
& 1.097 & \gc 1.118
& 1.028 & \gc 1.045
\\
\midrule

% ==================== Hourly ====================
\multirow{3}{*}{Hourly}
& sMAPE
& 21.441 & \gc 20.112
& 18.845 & \gc 18.564
& 17.391 & \gc 17.260
& 18.890 & \gc 17.910
& 19.610 & \gc 19.073
& 17.146 & \gc 16.904
\\
& MASE
& 3.578 & \gc 3.471
& 2.383 & \gc 2.144
& 1.772 & \gc 1.837
& 2.410 & \gc 2.106
& 2.353 & \gc 2.125
& 1.910 & \gc 1.825
\\
& OWA
& 1.330 & \gc 1.272
& 1.010 & \gc 0.952
& 0.843 & \gc 0.853
& 1.017 & \gc 0.927
& 1.025 & \gc 0.962
& 0.865 & \gc 0.841
\\
\midrule

% ==================== Weighted Average ====================
\multirow{3}{*}{\shortstack{Weighted\\Average}}
& sMAPE
& 13.479 & \gc \impv{\win{13.231}}{1.84\%}
& 12.402 & \gc \impv{\win{12.349}}{0.43\%}
& 12.554 & \gc \impv{\win{12.513}}{0.33\%}
& 13.647 & \gc \impv{\win{13.413}}{1.72\%}
& 13.096 & \gc \impv{\win{12.845}}{1.91\%}
& 12.893 & \gc \impv{\win{12.716}}{1.37\%}
\\

& MASE
& 1.864 & \gc \impv{\win{1.822}}{2.21\%}
& 1.676 & \gc \impv{\win{1.662}}{0.84\%}
& 1.680 & \gc \impv{\win{1.676}}{0.24\%}
& 1.873 & \gc \impv{\win{1.828}}{2.41\%}
& 1.729 & \gc \impv{\win{1.695}}{1.95\%}
& 1.718 & \gc \impv{\win{1.682}}{2.11\%}
\\

& OWA
& 0.984 & \gc \impv{\win{0.964}}{2.02\%}
& 0.895 & \gc \impv{\win{0.890}}{0.63\%}
& 0.902 & \gc \impv{\win{0.900}}{0.28\%}
& 0.993 & \gc \impv{\win{0.973}}{2.06\%}
& 0.935 & \gc \impv{\win{0.917}}{1.93\%}
& 0.925 & \gc \impv{\win{0.909}}{1.73\%}
\\
\bottomrule
\end{tabular}
}
\end{table*}
As shown in Table~\ref{tab:full_short_results}, \method improves all three weighted-average metrics across all six backbones. For Crossformer, the weighted-average sMAPE, MASE, and OWA decrease by 1.84\%, 2.21\%, and 2.02\%, respectively. On the Daily subset, \method reduces all three metrics for Crossformer by more than 10\%.

\subsection{Classification Results}
\paragraph{Setups.}
We follow the classification protocol described in Experimental Details. Each raw/\method-augmented pair uses the same classification head design, and results are averaged over three runs with seeds $42$, $43$, and $44$.

\paragraph{Results.}
\begin{table*}[thbp]
\centering
\newcommand{\win}[1]{\textcolor{red}{\textbf{#1}}}
\newcommand{\secbest}[1]{{#1}}

\caption{
Full classification results on 10 UEA datasets with and without
\method averaged over three seeds. The better result in each Raw/+\method pair is highlighted in \textcolor{red}{\textbf{bold red}}.
}
\label{tab:full_classification_results}

\setlength{\tabcolsep}{4.5pt}
\renewcommand{\arraystretch}{1.05}

\resizebox{\textwidth}{!}{%
\begin{tabular}{l|cc|cc|cc|cc|cc|cc}
\toprule
\multirow{2}{*}{\textbf{Dataset}}
& \multicolumn{2}{c|}{Crossformer \citeyearpar{crossformer}}
& \multicolumn{2}{c|}{PatchTST \citeyearpar{patchtst}}
& \multicolumn{2}{c|}{iTransformer \citeyearpar{itransformer}}
& \multicolumn{2}{c|}{TimeMixer \citeyearpar{timemixer}}
& \multicolumn{2}{c|}{TimeFilter \citeyearpar{timefilter}}
& \multicolumn{2}{c}{L-Drive \citeyearpar{ldrive}} \\
\cmidrule(lr){2-3}
\cmidrule(lr){4-5}
\cmidrule(lr){6-7}
\cmidrule(lr){8-9}
\cmidrule(lr){10-11}
\cmidrule(lr){12-13}
& Raw & +\method
& Raw & +\method
& Raw & +\method
& Raw & +\method
& Raw & +\method
& Raw & +\method \\
\midrule

Handwriting
& \secbest{18.98} & \win{19.02}
& \secbest{19.84} & \win{21.14}
& \secbest{11.76} & \win{18.35}
& \secbest{4.59}  & \win{5.45}
& \secbest{9.45}  & \win{10.67}
& \secbest{9.22}  & \win{9.45} \\

LSST
& \secbest{51.55} & \win{53.61}
& \win{54.07}     & \secbest{53.73}
& \secbest{57.91} & \win{58.57}
& \secbest{56.27} & \win{56.34}
& \secbest{60.34} & \win{60.84}
& \secbest{54.41} & \win{55.08} \\

NATOPS
& \secbest{73.33} & \win{75.00}
& \secbest{74.07} & \win{75.00}
& \secbest{80.19} & \win{80.37}
& \secbest{31.11} & \win{43.52}
& \secbest{77.96} & \win{78.70}
& \secbest{29.26} & \win{38.15} \\

RacketSports
& \secbest{71.93} & \win{73.03}
& \secbest{63.38} & \win{66.23}
& \secbest{70.18} & \win{76.97}
& \secbest{50.44} & \win{51.10}
& \secbest{61.62} & \win{63.16}
& \secbest{58.55} & \win{60.09} \\

SelfRegulationSCP2
& \secbest{55.00} & \win{55.19}
& \secbest{48.15} & \win{49.07}
& \secbest{51.30} & \win{55.74}
& \secbest{48.89} & \win{49.26}
& \win{51.85}     & \secbest{50.56}
& \secbest{50.56} & \win{51.11} \\

CharacterTrajectories
& \secbest{98.31} & \win{98.44}
& \win{95.66}     & \secbest{95.59}
& \secbest{97.63} & \win{97.66}
& \secbest{93.13} & \win{93.31}
& \secbest{97.03} & \win{97.28}
& \secbest{90.37} & \win{90.95} \\

SelfRegulationSCP1
& \secbest{85.21} & \win{86.01}
& \win{84.41}     & \secbest{84.30}
& \secbest{87.94} & \win{88.05}
& \secbest{77.02} & \win{79.18}
& \secbest{83.96} & \win{84.30}
& \win{71.10}     & \secbest{69.85} \\

ArticularyWordRecognition
& \secbest{85.11} & \win{85.22}
& \win{93.78}     & \win{93.78}
& \secbest{96.11} & \win{96.89}
& \secbest{26.11} & \win{31.00}
& \secbest{96.44} & \win{96.78}
& \secbest{37.67} & \win{39.89} \\

JapaneseVowels
& \secbest{72.70} & \win{76.49}
& \secbest{67.30} & \win{67.39}
& \secbest{94.77} & \win{95.59}
& \secbest{27.39} & \win{62.43}
& \secbest{69.73} & \win{72.25}
& \secbest{36.49} & \win{49.19} \\

UWaveGestureLibrary
& \secbest{64.38} & \win{65.21}
& \secbest{82.40} & \win{82.50}
& \secbest{80.62} & \win{81.15}
& \secbest{39.17} & \win{59.58}
& \secbest{83.23} & \win{86.15}
& \secbest{22.71} & \win{23.65} \\

\midrule
\textbf{Average Accuracy}
& \secbest{67.65} & \win{68.72}
& \secbest{68.31} & \win{68.87}
& \secbest{72.84} & \win{74.93}
& \secbest{45.41} & \win{53.12}
& \secbest{69.16} & \win{70.07}
& \secbest{46.03} & \win{48.74} \\

% \rowcolor{gray!10}
% \textbf{Avg. $\Delta$ (pp)}
% & \multicolumn{2}{c|}{\textbf{+1.07}}
% & \multicolumn{2}{c|}{\textbf{+0.57}}
% & \multicolumn{2}{c|}{\textbf{+2.09}}
% & \multicolumn{2}{c|}{\textbf{+7.71}}
% & \multicolumn{2}{c|}{\textbf{+0.91}}
% & \multicolumn{2}{c}{\textbf{+2.71}} \\

\bottomrule
\end{tabular}%
}
\end{table*}
As shown in Table~\ref{tab:full_classification_results}, \method improves classification accuracy in 54 out of 60 backbone--dataset combinations, with one tie. All six backbones achieve higher average accuracy after incorporating \method, with gains ranging from $0.57$ to $7.71$ percentage points. The largest improvement is for TimeMixer, while consistent gains are also obtained for stronger classifiers such as iTransformer and TimeFilter. These results show that \method improves average classification accuracy across all six backbones, including classifiers with strong baseline performance.

\subsection{Anomaly Detection Results}
\paragraph{Setups.}
We adopt a forecasting-based unsupervised detection protocol
\citep{msl_and_smap}, with implementation and training details
provided in Experimental Details.
Given a window ending at time $t$, the model predicts the next
observation $\mathbf{o}\in\mathbb{R}^{N}$.
The point-wise anomaly score is
\begin{equation}
s_{t+1}
=
\frac{1}{N}
\left\|
\mathbf{x}_{t+1}-\mathbf{o}
\right\|_2^2.
\end{equation}
Following prior high-quantile thresholding practices
\citep{horak2022nlp,cao2026pinfdit}, we set the detection
threshold $\tau$ to the 99.5th percentile of the normal
validation scores. A test point is identified as anomalous
when $s_{t+1}>\tau$. Test anomaly labels are used only for
final evaluation. We report Precision, Recall, and F1-score
averaged over the three runs.

\paragraph{Results.}
\method improves the average F1-score of every backbone. The largest average gain occurs on TimeFilter, from $82.73\%$ to $83.85\%$. Improvements are particularly pronounced on SMD: Crossformer and TimeFilter gain $5.50$ and $4.56$ percentage points in F1, respectively. In both cases, \method substantially increases precision while largely preserving recall, indicating that it reduces false alarms without substantially compromising anomaly sensitivity.
\begin{table*}[thbp]
\centering
\newcommand{\win}[1]{\textcolor{red}{\textbf{#1}}}
\newcommand{\secbest}[1]{{#1}}

\caption{
Full anomaly detection results on five datasets with and
without \method, averaged over three seeds. The better result in each
Raw/+\method pair is highlighted in \textcolor{red}{\textbf{bold red}}.
}
\label{tab:full_anomaly_results}

\setlength{\tabcolsep}{3.8pt}
\renewcommand{\arraystretch}{1.03}

\resizebox{\textwidth}{!}{%
\begin{tabular}{lc|cc|cc|cc|cc|cc|cc}
\toprule
\multirow{2}{*}{\textbf{Dataset}}
& \multirow{2}{*}{\textbf{Metric}}
& \multicolumn{2}{c|}{Crossformer \citeyearpar{crossformer}}
& \multicolumn{2}{c|}{PatchTST \citeyearpar{patchtst}}
& \multicolumn{2}{c|}{iTransformer \citeyearpar{itransformer}}
& \multicolumn{2}{c|}{TimeMixer \citeyearpar{timemixer}}
& \multicolumn{2}{c|}{TimeFilter \citeyearpar{timefilter}}
& \multicolumn{2}{c}{L-Drive \citeyearpar{ldrive}} \\
\cmidrule(lr){3-4}\cmidrule(lr){5-6}\cmidrule(lr){7-8}
\cmidrule(lr){9-10}\cmidrule(lr){11-12}\cmidrule(lr){13-14}
& & Raw & +\method & Raw & +\method & Raw & +\method
  & Raw & +\method & Raw & +\method & Raw & +\method \\
\midrule

\multirow{3}{*}{SMD}
& Precision & 56.08 & 64.26 & 70.92 & 71.27 & 72.72 & 74.46 & 66.52 & 67.56 & 63.14 & 70.91 & 70.87 & 70.98 \\
& Recall    & 88.47 & 87.70 & 89.22 & 88.33 & 87.84 & 88.01 & 89.85 & 89.98 & 89.05 & 87.70 & 88.38 & 88.43 \\
& F1
& \secbest{68.64} & \win{74.14}
& \win{79.03} & \secbest{78.89}
& \secbest{79.56} & \win{80.67}
& \secbest{76.44} & \win{77.13}
& \secbest{73.81} & \win{78.37}
& \secbest{78.65} & \win{78.75} \\
\midrule

\multirow{3}{*}{MSL}
& Precision & 82.38 & 82.43 & 80.58 & 79.96 & 79.75 & 79.86 & 81.15 & 80.97 & 80.17 & 81.89 & 81.27 & 80.25 \\
& Recall    & 87.91 & 87.91 & 87.27 & 88.22 & 88.22 & 88.22 & 87.27 & 88.58 & 87.74 & 87.63 & 87.27 & 87.74 \\
& F1
& \secbest{85.05} & \win{85.08}
& \secbest{83.79} & \win{83.89}
& \secbest{83.77} & \win{83.83}
& \secbest{84.10} & \win{84.59}
& \secbest{83.77} & \win{84.62}
& \win{84.16} & \secbest{83.82} \\
\midrule

\multirow{3}{*}{SMAP}
& Precision & 92.98 & 93.48 & 94.20 & 94.13 & 93.34 & 91.80 & 93.78 & 93.83 & 93.70 & 95.04 & 93.99 & 94.39 \\
& Recall    & 55.83 & 55.76 & 55.41 & 55.41 & 55.41 & 55.81 & 55.41 & 55.41 & 55.41 & 54.93 & 55.41 & 55.41 \\
& F1
& \secbest{69.76} & \win{69.85}
& \win{69.78} & \secbest{69.76}
& \win{69.54} & \secbest{69.41}
& \secbest{69.66} & \win{69.68}
& \win{69.64} & \secbest{69.61}
& \secbest{69.72} & \win{69.83} \\
\midrule

\multirow{3}{*}{PSM}
& Precision & 99.68 & 99.69 & 99.56 & 99.56 & 99.55 & 99.53 & 99.43 & 99.44 & 99.52 & 99.51 & 99.46 & 99.40 \\
& Recall    & 84.94 & 84.96 & 86.78 & 86.95 & 85.31 & 85.66 & 90.07 & 88.64 & 85.15 & 85.15 & 87.27 & 88.68 \\
& F1
& \secbest{91.72} & \win{91.73}
& \secbest{92.72} & \win{92.82}
& \secbest{91.88} & \win{92.08}
& \win{94.52} & \secbest{93.72}
& \win{91.78} & \secbest{91.77}
& \secbest{92.96} & \win{93.72} \\
\midrule

\multirow{3}{*}{SWaT}
& Precision & 93.42 & 90.09 & 94.54 & 94.65 & 95.22 & 95.36 & 94.58 & 94.71 & 93.53 & 93.96 & 94.58 & 94.78 \\
& Recall    & 95.64 & 94.81 & 95.81 & 95.81 & 95.81 & 95.81 & 95.81 & 95.81 & 95.81 & 95.81 & 95.81 & 95.81 \\
& F1
& \win{94.52} & \secbest{92.24}
& \secbest{95.17} & \win{95.23}
& \secbest{95.52} & \win{95.58}
& \secbest{95.19} & \win{95.26}
& \secbest{94.66} & \win{94.88}
& \secbest{95.19} & \win{95.30} \\
\midrule

\multicolumn{2}{c|}{\textbf{Average F1}}
& \secbest{81.94} & \win{82.61}
& \secbest{84.10} & \win{84.12}
& \secbest{84.05} & \win{84.31}
& \secbest{83.98} & \win{84.07}
& \secbest{82.73} & \win{83.85}
& \secbest{84.14} & \win{84.28} \\
\bottomrule
\end{tabular}%
}
\end{table*}

\subsection{Ablation Study}
\label{app:full_ablation}

\paragraph{Setups.}
We conduct the ablation with TimeFilter on the Noisy Lorenz63 datasets at
$\mathrm{SNR}\in\{3,5,7\}$, using an input length of 96 and prediction horizons
of 16 and 64. We compare six configurations: Raw, Raw + RC,
Analyt. + [Amp., Freq., $O$], Analyt. + $\phi$,
Analyt. + [Amp., Freq., $O$] + RC, and the full \method.
Raw + RC applies the same RC branch directly to the raw input without
analyticization, controlling for the additional prediction branch. Analyt.
denotes the analytic IMFs obtained through reflection-padded EMD and Hilbert
analysis. The gate uses either the directly computed amplitude, frequency, and
IMF order, or the four analytic features $\phi$. All variants share the same data splits, TimeFilter configuration, training protocol, and seed. For each dataset--horizon pair, we reuse the corresponding preselected hyperparameters without retuning on the test set. We select checkpoints by validation MAE and report MSE and MAE from the same selected checkpoint. Table~\ref{tab:lorenz_ablation} averages the two horizons, while the complete horizon-level results are reported below. 

\paragraph{Results.}
\begin{table*}[thbp]
  \centering
  \caption{\textbf{Detailed ablation results on Noisy Lorenz63.}
  Each entry reports MSE / MAE for TimeFilter at seed 42.
  Analyt. denotes the analytic IMFs.}
  \label{tab:full_lorenz_ablation}
  \resizebox{\textwidth}{!}{%
    \begin{tabular}{llcccccc}
      \toprule
      Dataset & Horizon
      & Raw
      & Raw + RC
      & Analyt. + [Amp., Freq., $O$]
      & Analyt. + $\phi$
      & Analyt. + [Amp., Freq., $O$] + RC
      & \method (Ours) \\
      \midrule

      SNR=3 & 16
      & 0.551 / 0.391
      & 0.578 / 0.426
      & 0.582 / 0.427
      & 0.546 / 0.389
      & 0.524 / 0.378
      & \textcolor{red}{\textbf{0.508 / 0.363}} \\

      SNR=3 & 64
      & 0.848 / 0.620
      & 0.849 / 0.628
      & 0.843 / 0.617
      & 0.886 / 0.651
      & 0.880 / 0.645
      & \textcolor{red}{\textbf{0.839 / 0.616}} \\
      \midrule

      SNR=5 & 16
      & 0.508 / 0.415
      & 0.451 / 0.369
      & 0.499 / 0.408
      & 0.490 / 0.399
      & 0.412 / 0.333
      & \textcolor{red}{\textbf{0.411 / 0.332}} \\

      SNR=5 & 64
      & 0.788 / 0.611
      & 0.821 / 0.635
      & 0.827 / 0.635
      & 0.843 / 0.652
      & 0.792 / 0.615
      & \textcolor{red}{\textbf{0.778 / 0.608}} \\
      \midrule

      SNR=7 & 16
      & 0.414 / 0.377
      & 0.329 / 0.298
      & 0.420 / 0.380
      & 0.363 / 0.334
      & 0.324 / 0.295
      & \textcolor{red}{\textbf{0.320 / 0.289}} \\

      SNR=7 & 64
      & 0.795 / 0.640
      & 0.794 / 0.641
      & 0.781 / 0.621
      & 0.733 / 0.594
      & 0.734 / 0.596
      & \textcolor{red}{\textbf{0.724 / 0.586}} \\

      \bottomrule
    \end{tabular}%
  }
\end{table*}
Table~\ref{tab:full_lorenz_ablation} shows that \method achieves the lowest MSE and MAE in all six SNR--horizon settings. Raw + RC is not consistently beneficial: it improves the short-horizon cases at SNR=5 and 7, but degrades both SNR=3 settings and the long-horizon SNR=5 case. The no-RC analytic variants are similarly inconsistent: neither they nor the Raw + RC control match the full model across the six evaluated settings. Compared with its no-RC counterpart, adding RC to the $\phi$-based variant improves both metrics in all six settings. The full model also outperforms the Raw + RC and direct-attribute RC controls at both horizons for every SNR. These results are consistent with a complementary interaction between analytic feature encoding and residual-aware correction, with $\phi$ providing a stronger basis for correction than the direct attributes.

\end{document}